\documentclass{article} 
\usepackage{iclr2025_conference,times}

\usepackage{url}
\usepackage{booktabs}
\usepackage{amsfonts}
\usepackage{nicefrac}
\usepackage{longtable}
\usepackage{microtype}
\usepackage{colortbl}

\usepackage{amsmath}
\usepackage{amssymb}
\usepackage{caption}
\usepackage{enumitem}
\usepackage{float}
\usepackage{multirow}
\usepackage{tcolorbox}
\usepackage{xcolor}

\definecolor{thepurple}{rgb}{0.5882352941176471, 0.45098039215686275, 0.6509803921568628}
\definecolor{citecolor}{rgb}{0.0, 0.2, 0.45}
\definecolor{linkcolor}{rgb}{0.55, 0.0, 0.35}
\definecolor{urlcolor}{rgb}{0.0, 0.2, 0.55}

\definecolor{fontmlp}{HTML}{185D8B}
\definecolor{fontcollective}{HTML}{2A8424}
\definecolor{fontindividual}{HTML}{B71517}

\usepackage[pagebackref=true,breaklinks=true,colorlinks,citecolor=citecolor,linkcolor=linkcolor,urlcolor=urlcolor,bookmarks=false]{hyperref}
\usepackage{graphicx}
\usepackage[inkscapelatex=false]{svg}

\usepackage{lib}

\title{\model: Advancing Tabular Deep Learning\\ with Parameter-Efficient Ensembling}

\author{
Yury Gorishniy \thanks{The corresponding author: \texttt{yurygorishniy@gmail.com}} \\
Yandex
\And
Akim Kotelnikov \\
HSE University, Yandex
\And
Artem Babenko \\
Yandex
}

\iclrfinalcopy 
\begin{document}

\maketitle

\begin{abstract}
Deep learning architectures for supervised learning on tabular data range from simple multilayer perceptrons (MLP) to sophisticated Transformers and retrieval-augmented methods.
This study highlights a major, yet so far overlooked opportunity for designing substantially better MLP-based tabular architectures.
Namely, our new model TabM relies on \textit{efficient ensembling}, where one TabM efficiently imitates an ensemble of MLPs and produces multiple predictions per object.
Compared to a traditional deep ensemble, in TabM, the underlying implicit MLPs are trained simultaneously, and (by default) share most of their parameters, which results in significantly better performance and efficiency.
Using TabM as a new baseline, we perform a large-scale evaluation of tabular DL architectures on public benchmarks in terms of both task performance and efficiency, which renders the landscape of tabular DL in a new light.
Generally, we show that MLPs, including \model, form a line of stronger and more practical models compared to attention- and retrieval-based architectures.
In particular, we find that \model\ demonstrates the best performance among tabular DL models.
Then, we conduct an empirical analysis on the ensemble-like nature of \model.
We observe that the multiple predictions of \model\ are weak individually, but powerful collectively.
Overall, our work brings an impactful technique to tabular DL and advances the performance-efficiency \mbox{trade-off} with \model\ --- a simple and powerful baseline for researchers and practitioners.
The code is available at: \url{\repository}.
\end{abstract}

\section{Introduction}

Supervised learning on tabular data is a ubiquitous machine learning (ML) scenario in a wide range of industrial applications.
Among classic non-deep-learning methods, the state-of-the-art solution for such tasks is gradient-boosted decision trees (GBDT) \citep{prokhorenkova2018catboost, chen2016xgboost, ke2017lightgbm}.
Deep learning (DL) models for tabular data, in turn, are reportedly improving, and the most recent works claim to perform on par or even outperform GBDT on academic benchmarks \citep{hollmann2022tabpfn, chen2023trompt, chen2023excelformer, gorishniy2023tabr}.

However, from the practical perspective, it is unclear if tabular DL offers any obvious go-to baselines beyond simple architectures in the spirit of a multilayer perceptron (MLP).
\textit{First}, the scale and consistency of performance improvements of new methods w.r.t. simple MLP-like baselines are not always explicitly analyzed in the literature.
Thus, one has to infer those statistics from numerous per-dataset performance scores, which makes it hard to reason about the progress.
At the same time, due to the extreme diversity of tabular datasets, consistency is an especially valuable and hard-to-achieve property for a hypothetical go-to baseline.
\textit{Second}, efficiency-related properties, such as training time, and especially inference throughput, sometimes receive less attention.
While methods are usually equally affordable on small-to-medium datasets (e.g. $<$100K objects), their applicability to larger datasets remains uncertain.
\textit{Third}, some recent work generally suggests that the progress on academic benchmarks may not transfer that well to real-world tasks \citep{rubachev2024tabred}.
With all the above in mind, in this work, we thoroughly evaluate existing tabular DL methods and find that non-MLP models do not yet offer a convincing replacement for MLPs.

At the same time, we identify a previously overlooked path towards more powerful, reliable, and reasonably efficient tabular DL models.
In a nutshell, we find that the parameter-efficient approach to deep ensembling, where most weights are shared between ensemble members, allow one to make simple and strong tabular models out of plain MLPs.
For example, MLP coupled with BatchEnsemble \citep{wen2020batchensemble} --- a long-existing method --- right away outperforms popular attention-based models, such as FT-Transformer \citep{gorishniy2021revisiting}, while being simpler and more efficient.
This result alone suggests that efficient ensembling is a low-hanging fruit for tabular DL.

Our work builds on the above observations and offers \model\ --- a new powerful and practical model for researchers and practitioners.
Drawing an informal parallel with GBDT (an ensemble of decision trees), \model\ can also be viewed as a simple base model (MLP) combined with an ensembling-like technique, providing high performance and simple implementation at the same time.

\textbf{Main contributions.}
We summarize our main contributions as follows:

\begin{enumerate}[nosep,leftmargin=2em]

    \item
    We present \model\ --- a simple DL architecture for supervised learning on tabular data.
    \model\ is based on MLP and parameter-efficient ensembling techniques closely related to BatchEnsemble \citep{wen2020batchensemble}.
    In particular, \model\ produces \textbf{M}ultiple predictions per object.
    \model\ easily competes with GBDT and outperforms prior tabular DL models, while being more efficient than attention- and retrieval-based DL architectures.

    \item
    We provide a fresh perspective on tabular DL models in a large-scale evaluation along four dimensions: performance ranks, performance score distributions, training time, and inference throughput.
    One of our findings is that MLPs, including \model, hit an appealing performance-efficiency tradeoff, which is not the case for attention- and retrieval-based models.

    \item
    We show that the two key reasons for TabM's high performance are the collective training of the underlying implicit MLPs and the weight sharing.
    We also show that the multiple predictions of \model\ are weak and overfitted individually, while their average is strong and generalizable.

\end{enumerate}

\section{Related work}
\label{sec:related-work}

\textbf{Decision-tree-based models.}
Gradient-boosted decision trees (GBDT) \citep{chen2016xgboost, ke2017lightgbm, prokhorenkova2018catboost} is a strong and efficient baseline for tabular tasks.
GBDT is a classic machine learning model, specifically, an ensemble of decision trees.
Our model \model\ is a deep learning model, specifically, a parameter-efficient ensemble of MLPs.

\textbf{Tabular deep learning architectures.}
A large number of deep learning architectures for tabular data have been proposed over the recent years.
That includes attention-based architectures \citep{song2019autoint,gorishniy2021revisiting,somepalli2021saint,kossen2021self,yan2023t2g}, retrieval-augmented architectures \citep{somepalli2021saint,kossen2021self,gorishniy2023tabr,ye2024modern}, MLP-like models \citep{gorishniy2021revisiting,klambauer2017self,wang2020dcn2} and others \citep{arik2020tabnet,popov2020neural,chen2023trompt,marton2024grande,hollmann2022tabpfn}.
Compared to prior work, the key difference of our model \model\ is its computation flow, where one \model\ imitates an ensemble of MLPs by producing multiple independently trained predictions.
Prior attempts to bring ensemble-like elements to tabular DL \citep{badirli2020gradient,popov2020neural} were not found promising \citep{gorishniy2021revisiting}.
Also, being a simple feed-forward MLP-based model, \model\ is significantly more efficient than some of the prior work.
Compared to attention-based models, \model\ does not suffer from quadratic computational complexity w.r.t. the dataset dimensions.
Compared to retrieval-based models, \model\ is easily applicable to large datasets.

\textbf{Improving tabular MLP-like models.}
Multiple recent studies achieved competitive performance with MLP-like architectures on tabular tasks by applying architectural modifications \citep{gorishniy2022embeddings}, regularizations \citep{kadra2021well,jeffares2023tangos,holzmüller2024better}, custom training techniques \citep{bahri2021scarf,rubachev2022revisiting}.
Thus, it seems that tabular MLPs have good potential, but one has to deal with overfitting and optimization issues to reveal that potential.
Our model \model\ achieves high performance with MLP in a different way, namely, by using it as the base backbone in a parameter-efficient ensemble in the spirit of BatchEsnsemble \citep{wen2020batchensemble}.
Our approach is orthogonal to the aforementioned training techniques and architectural advances.

\textbf{Deep ensembles.}
In this paper, by a deep ensemble, we imply multiple DL models of the same architecture trained independently \citep{jeffares2023joint} for the same task under different random seeds (i.e. with different initializations, training batch sequences, etc.).
The prediction of a deep ensemble is the mean prediction of its members.
Deep ensembles often significantly outperform single DL models of the same architecture \citep{fort2020deep} and can excel in other tasks like uncertainty estimation or out-of-distribution detection \citep{lakshminarayanan2017simple}.
It was observed that individual members of deep ensembles can learn to extract diverse information from the input, and the power of deep ensembles depends on this diversity \citep{allenzhu2023towards}.
The main drawback of deep ensembles is the cost and inconvenience of training and using multiple models.

\textbf{Parameter-efficient deep ``ensembles''.}
To achieve the performance of deep ensembles at a lower cost, multiple studies proposed architectures that imitate ensembles by producing multiple predictions with one model \citep{lee2015why,zhang2020diversified,wen2020batchensemble,havasi2020training,antoran2020depth,turkoglu2022film}.
Such models can be viewed as ``ensembles'' where the implicit ensemble members share a large amount of their weights.
There are also non-architectural approaches to efficient ensembling, e.g. FGE \citep{garipov2018loss}, but we do not explore them, because we are interested specifically in architectural techniques.
In this paper, we highlight parameter-efficient ensembling as an impactful paradigm for tabular DL.
In particular, we describe two simple variations of BatchEnsemble \citep{wen2020batchensemble} that are highly effective for tabular MLPs.
One variation uses a more efficient parametrization, and another one uses an improved initialization.

\section{\model}
\label{sec:model}

In this section, we present \model\ --- a \textbf{Tab}ular DL model that makes \textbf{M}ultiple predictions.

\subsection{Preliminaries}
\label{sec:model-preliminaries}

\textbf{Notation.}
We consider classification and regression tasks on tabular data.
$x$ and $y$ denote the features and a label, respectively, of one object from a given dataset.
A machine learning model takes $x$ as input and produces $\hat{y}$ as a prediction of $y$.
$N \in \mathbb{N}$ and $d \in \mathbb{N}$ respectively denote the ``depth'' (e.g. the number of blocks) and ``width'' (e.g. the size of the latent representation) of a given neural network.
$d_y \in \mathbb{N}$ is the output representation size (e.g. $d_y = 1$ for regression tasks, and $d_y$ equals the number of classes for classification tasks).

\textbf{Datasets.}
Our benchmark consists of \ndatasets\ publicly available datasets used in prior work, including \citet{grinsztajn2022why,gorishniy2023tabr,rubachev2024tabred}.
The main properties of our benchmark are summarized in \autoref{tab:datasets}, and more details are provided in \autoref{A:sec:datasets}.

\begin{table}[h!]
    \centering
    \setlength\tabcolsep{5pt}
    \caption{
        The overview of our benchmark.
        The ``Split type'' property is explained in the text.
    }
    \scalebox{0.8}{    \begin{tabular}{ccccccccccccc}
    \toprule
    \multicolumn{1}{c}{\#Datasets} &
    \multicolumn{4}{c}{Train size} &
    \multicolumn{4}{c}{\#Features} & 
    \multicolumn{2}{c}{Task type} &
    \multicolumn{2}{c}{Split type}
    \\
    \cmidrule(lr){1-1} \cmidrule(lr){2-5} \cmidrule(lr){6-9} \cmidrule(lr){10-11} \cmidrule(lr){12-13}
    & {\footnotesize Min.} & {\footnotesize Q50}& {\footnotesize Mean} & {\footnotesize Max.} & {\footnotesize Min.} & {\footnotesize Q50}& {\footnotesize Mean} & {\footnotesize Max.} & {\footnotesize \#Regr.} & {\footnotesize \#Classif.} & {\footnotesize Random} & {\footnotesize Domain-aware}
    \\ 
    \midrule 
    \ndatasets & 1.8K & 12K & 76K & 723K & 3 & 20 & 108 & 986 & 28 & 18 & 37 & 9 \\
    \bottomrule 
    \end{tabular}}
    \label{tab:datasets}
\end{table}

\textbf{Domain-aware splits.}
We pay extra attention to datasets with what we call ``domain-aware'' splits, including the eight datasets from the TabReD benchmark \citep{rubachev2024tabred} and the Microsoft dataset \citep{microsoft}.
For these datasets, their original real-world splits are available, e.g. time-aware splits as in TabReD.
Such datasets were shown to be challenging for some methods because they naturally exhibit a certain degree of distribution shift between training and test parts \citep{rubachev2024tabred}.
The random splits of the remaining \nrandomsplits\ datasets are inherited from prior work.

\textbf{Experiment setup.}
We use the setup from \cite{gorishniy2023tabr}, and describe it in detail in \autoref{A:sec:impl-experiment-setup}.
Most importantly, on each dataset, a given model undergoes hyperparameter tuning on the \textit{validation} set, then the tuned model is trained from scratch under multiple random seeds, and the \textit{test} metric averaged over the random seeds becomes the final score of the model on the dataset.

\textbf{Metrics.}
We use RMSE (the root mean square error) for regression tasks, and accuracy or ROC-AUC for classification tasks depending on the dataset source.
See \autoref{A:sec:impl-metrics} for details.

Also, throughout the paper, we often use the relative performance of models w.r.t. MLP as the key metric.
This metric gives a unified perspective on all tasks and allows reasoning about the scale of improvements w.r.t. to a simple baseline (MLP).
Formally, on a given dataset, the metric is defined as $\left( \frac{\text{score}}{\text{baseline}} - 1\right) \cdot 100\%$, where ``score'' is the metric of a given model, and ``baseline'' is the metric of MLP.
In this computation, for regression tasks, we convert the raw metrics from RMSE to $R^2$ to better align the scales of classification and regression metrics.

\subsection{A quick introduction to BatchEnsemble.}
\label{sec:model-batchensemble}

For a given architecture, let's consider any linear layer $l$ in it: $l(x) = Wx + b$, where $x \in \R^{d_1}$, $W \in \R^{d_2 \times d_1}$, $b \in \R^{d_2}$.
To simplify the notation, let $d_1 = d_2 = d$.
In a traditional deep ensemble, the $i$-th member has its own set of weights $W_i, b_i$ for this linear layer: \mbox{$l_i(x_i) = W_ix_i + b_i$}, where $x_i$ is the object representation within the $i$-th member.
By contrast, in BatchEnsemble, this linear layer is either (1) fully shared between all members, or (2) mostly shared: \mbox{$l_i(x_i) = s_i \odot (W(r_i \odot x_i)) + b_i$}, where $\odot$ is the elementwise multiplication, $W \in \R^{d \times d}$ is shared between all members, and $r_i, s_i, b_i \in \R^d$ are \textit{not} shared between the members.
This is equivalent to defining the $i$-th weight matrix as \mbox{$W_i = W \odot (s_ir_i^T)$}.
To ensure diversity of the ensemble members, $r_i$ and $s_i$ of all members are initialized randomly with $\pm 1$.
All other layers are fully shared between the members of BatchEnsemble.

The described parametrization allows packing all ensemble members in one model that simultaneously takes $k$ objects as input, and applies all $k$ implicit members in parallel, without explicitly materializing each member.
This is achieved by replacing one or more linear layers of the original neural network with their BatchEnsemble versions:
\mbox{$l_\text{BE}(X) = ((X \odot R) W) \odot S + B$},
where \mbox{$X \in \R^{k \times d}$} stores $k$ object representations (one per member), and $R, S, B \in \R^d$ store the non-shared weights ($r_i$, $s_i$, $b_i$) of the members, as shown at the lower left part of \autoref{fig:model}.

\textbf{Terminology.}
In this paper, we call $r_i$, $s_i$, $b_i$, $R$, $S$ and $B$ \textit{adapters}, and the implicit members of parameter-efficient emsembles (e.g. BatchEnsemble) --- \textit{implicit submodels} or simply \textit{submodels}.
\\\textbf{Overhead to the model size.}
With BatchEnsemble, adding a new ensemble member means adding only one row to each of the matrices $R$, $S$, and $B$, which results in $3d$ new parameters per layer.
For typical values of $d$, this is a negligible overhead to the original layer size $d^2 + d$.
\\\textbf{Overhead to the runtime.}
Thanks to the modern hardware, the large number of shared weights and the parallel execution of the $k$ forward passes, the runtime overhead of BatchEnsemble can be (significantly) lower than $\times k$ \citep{wen2020batchensemble}.
Intuitively, if the original workload underutilizes the hardware, there are more chances to pay less than $\times k$ overhead.

\subsection{Architecture}
\label{sec:model-design}

TabM is one model representing an ensemble of $k$ MLPs.
Contrary to conventional deep ensembles, in TabM, the $k$ MLPs are trained in parallel and share most of their weights by default, which leads to better performance and efficiency.
We present multiple variants of TabM that differ in their weight-sharing strategies, where \model\ and \modelmini\ are the most effective variants, and \modelpacked\ is a conceptually important variant potentially useful in some cases.
We obtain our models in several steps, starting from essential baselines.
We always use the ensemble size $k=32$ and analyze this hyperparameter in \autoref{sec:analysis-k}.
In \autoref{A:sec:model-motivation}, we explain that using MLP as the base model is crucial because of its excellent efficiency.

\textbf{MLP.}
We define MLP as a sequence of $N$ simple blocks followed by a linear prediction head:
\\$\text{MLP}(x) = \text{Linear}(\text{Block}_N(\ldots(\text{Block}_1(x)))$, where $\text{Block}_i(x) = \text{Dropout}(\text{ReLU}(\text{Linear}((x)))$.

\begin{figure}[h!]
    \centering
    \includegraphics[width=0.99\linewidth]{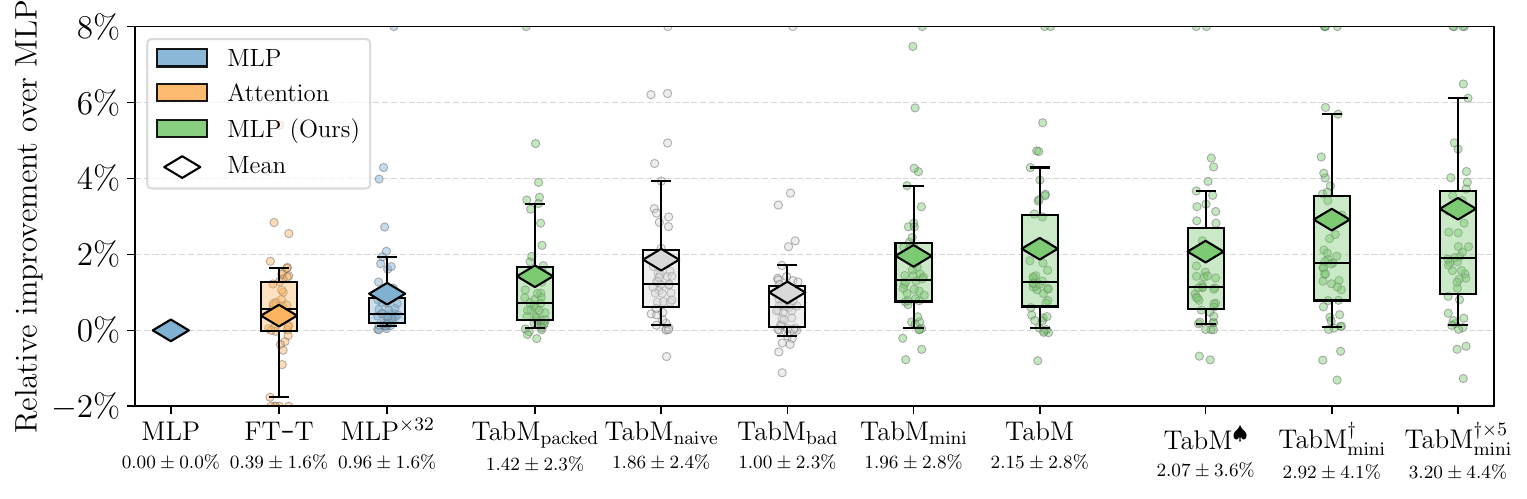}
    \caption{
        The performance of models described in \autoref{sec:model-design} on \ndatasets\ datasets from \autoref{tab:datasets}; plus several baselines on the left.
        For a given model, one dot on a jitter plot describes the performance score on one of the \ndatasets\ datasets.
        The box plots describe the percentiles of the jitter plots: the boxes describe the 25th, 50th, and 75th percentiles, and the whiskers describe the 10th and 90th percentiles.
        Outliers are clipped.
        The numbers at the bottom are the mean and standard deviations over the jitter plots.
        For each model, hyperparameters are tuned.
        ``$\text{Model}^{\times k}$'' denotes an ensemble of $k$ models.
    }
    \label{fig:model-design}
\end{figure}

\textbf{\mlpxk\ = MLP + Deep Ensemble.}
We denote the traditional deep ensemble of $k$ independently trained MLPs as \mlpxk.
To clarify, this means tuning hyperparameters of one MLP, then independently training $k$ tuned MLPs under different random seeds, and then averaging their predictions.
The performance of \mlpxk\ is reported in \autoref{fig:model-design}.
Notably, the results are already better and more stable than those of FT-Transformer \citep{gorishniy2021revisiting} --- the popular attention-based baseline.

Although the described approach is a somewhat default way to implement an ensemble, it is not optimized for the task performance of the ensemble.
First, for each of the $k$ MLPs, the training is stopped based on the individual validation score, which is optimal for each individual MLP, but can be suboptimal for their ensemble.
Second, the hyperparameters are also tuned for one MLP without knowing about the subsequent ensembling.
All TabM variants are free from these issues.

\begin{figure}[h!]
    \centering
    \includegraphics[width=0.99\linewidth]{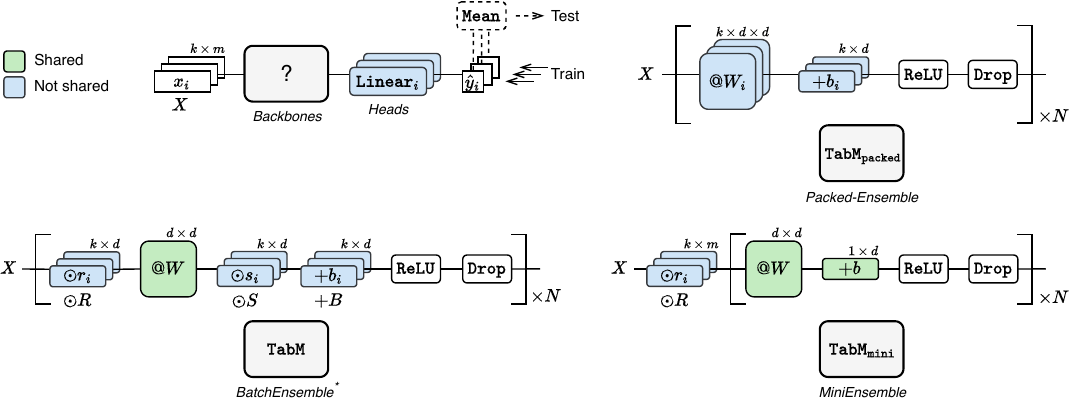}
    \caption{
        \textit{(Upper left)}
        A high-level illustration of TabM.
        One TabM represents an ensemble of $k$ MLPs processing $k$ inputs in parallel.
        The remaining parts of the figure are three different parametrizations of the $k$ MLP backbones.
        \textit{(Upper right)}
        \modelpacked\ consists of $k$ fully independent MLPs.
        \mbox{\textit{(Lower left)}}
        \model\ is obtained by injecting three non-shared adapters $R$, $S$, $B$ in each of the $N$ linear layers of \textit{one} MLP (\mbox{$^*$ the} initialization differs from \citet{wen2020batchensemble}).
        \textit{(Lower right)}
        \modelmini\ is obtained by keeping only the very first adapter $R$ of \model\, and removing the remaining $3N - 1$ adapters.
        \textit{(Details)}
        Input transformations such as one-hot-encoding or feature embeddings \citep{gorishniy2022embeddings} are omitted for simplicity.
        \texttt{Drop} denotes dropout \citep{srivastava2014dropout}.
    }
    \label{fig:model}
\end{figure}

\textbf{\modelpacked\ = MLP + Packed-Ensemble.}
As the first step towards better and more efficient ensembles of MLPs, we implement $k$ MLPs as one large model using Packed-Ensemble \citep{laurent2022packed}.
This results in \modelpacked\ illustrated in
\autoref{fig:model}.
As an architecture, \modelpacked\ is equivalent to \mlpxk\ and stores $k$ independent MLPs without any weight sharing.
However, the critical difference is that TabM
processes $k$ inputs in parallel, which means that one training step of TabM consists of $k$ parallel training steps of the individual MLPs.
This allows monitoring the performance of the ensemble during the training and stopping the training when it is optimal for the whole ensemble, not for individual MLPs.
As a consequence, this also allows tuning hyperparameters for \modelpacked\ as for one model.
As shown in \autoref{fig:model-design}, \modelpacked\ delivers significantly better performance compared to \mlpxk.
Efficiency-wise, for typical depth and width of MLPs, the runtime overhead of \modelpacked\ is noticeably less than $\times k$ due to the parallel execution of the $k$ forward passes on the modern hardware.
Nevertheless, the $\times k$ overhead of \modelpacked\ to the model size motivates further exploration.

\textbf{\modelnaive\ = MLP + BatchEnsemble.}
To reduce the size of \modelpacked, we now turn to weight sharing between the MLPs, and naively apply BatchEnsemble \citep{wen2020batchensemble} instead of Packed-Ensemble, as described in \autoref{sec:model-batchensemble}.
This gives us \modelnaive  --- a preliminary version of \model.
In fact, the architecture (but not the initialization) of \modelnaive\ is already equivalent to that of \model, so
\autoref{fig:model} is applicable.
Interestingly, \autoref{fig:model-design} reports higher performance of \modelnaive\ compared to \modelpacked.
Thus, constraining the ensemble with weight sharing turns out to be a highly effective regularization on tabular tasks.
The alternatives to BatchEnsemble are discussed in \autoref{A:sec:model-motivation}.

\textbf{\modelmini\ = MLP + MiniEnsemble.}
By construction, the just discussed \modelnaive\ (illustrated as ``\model'' in \autoref{fig:model}) has $3N$ adapters: $R$, $S$ and $B$ in each of the $N$ blocks.
Let's consider the very first adapter, i.e. the first adapter $R$ in the first linear layer.
Informally, its role can be described as mapping the $k$ inputs living in the same representation space to $k$ different representation spaces \textit{before} the tabular features are mixed with $@ W$ for the first time.
A simple experiment reveals that this adapter is critical.
First, we remove it from \modelnaive\ and keep the remaining $3N-1$ adapters untouched, which gives us \modelbad\ with worse performance, as shown in \autoref{fig:model-design}.
Then, we do the opposite: we keep only the very first adapter of \modelnaive\ and remove the remaining $3N - 1$ adapters, which gives us \modelmini\ --- the minimal version of \model.
\modelmini\ is illustrated in \autoref{fig:model}, where we call the described approach ``MiniEnsemble''.
\autoref{fig:model-design} shows that \modelmini\ performs even slightly better than \modelnaive, despite having only one adapter instead of $3N$ adapters.

\textbf{\model\ = MLP + BatchEnsemble + Better initialization.}
The just obtained results motivate the next step.
We go back to the architecture of \modelnaive\ with all $3N$ adapters, but initialize all multiplicative adapters $R$ and $S$, except for the very first one, deterministically with $1$.
As such, at initialization, the deterministically initialized adapters have no effect, and the model behaves like \modelmini, but these adapters are free to add more expressivity during training.
This gives us \model, illustrated in \autoref{fig:model}.
\autoref{fig:model-design} shows that \model\ is the best variation so far.

\textbf{Hyperparameters.}
Compared to MLP, the only new hyperparameter of \model\ is $k$ --- the number of implicit submodels.
We heuristically set $k=32$ and do not tune this value.
We analyze the influence of $k$ in \autoref{sec:analysis-k}.
We also share additional observations on the learning rate in \autoref{A:sec:model-hyperparameters}.

\textbf{Limitations and practical considerations} are commented in \autoref{A:sec:model-limitations}.

\subsection{Important practical modifications of \model}
\label{sec:model-important}

$\mathbf{\spadesuit \sim}$ \textbf{Shared training batches}.
Recall that the order of training objects usually varies between ensemble members, because of the random shuffling with different seeds.
For \model, in terms of \autoref{fig:model}, that corresponds to $X$ storing $k$ different training objects $\{x_i\}_{i = 1}^k$.
We observed that reusing the training batches between the \model's submodels results in only minor performance loss on average (depending on a dataset), as illustrated with \modelspade\ in \autoref{fig:model-design}.
In practice, due to the simpler implementation and better efficiency, sharing training batches can be a reasonable starting point.

$\mathbf{\dagger \sim}$ \textbf{Non-linear feature embeddings}.
In \autoref{fig:model-design}, \modelminiemb\ denotes \modelmini\ with non-linear feature embeddings from \citep{gorishniy2022embeddings}, which demonstrates the high utility of feature embeddings for \model.
Specifically, we use a slightly modified version of the piecewise-linear embeddings (see \autoref{A:sec:impl-feature-embeddings} for details).

$\mathbf{\times N \sim}$ \textbf{Deep ensemble}.
In \autoref{fig:model-design}, \modelminiembensfive\ denotes an ensemble of five independent \modelminiemb\ models, showing that \model\ itself can benefit from the conventional deep ensembling.

\subsection{Summary}
\label{sec:model-summary}

The story behind \model\ shows that technical details of \textit{how} to construct and train an ensemble have a major impact on task performance.
Most importantly, we highlight simultaneous training of the (implicit) ensemble members and weight sharing between them.
The former is responsible for the ensemble-aware stopping of the training, and the latter apparently serves as a form of regularization.

\section{Evaluating tabular deep learning architectures}
\label{sec:evaluation}

Now, we perform an empirical comparison of many tabular models, including \model.

\subsection{Baselines}
\label{sec:evaluation-baselines}

In the main text, we use the following baselines:
MLP (defined in \autoref{sec:model-design}), FT-Transformer denoted as ``FT-T'' (the attention-based model from \citet{gorishniy2021revisiting}), SAINT (the attention- and retrieval-based model from \citet{somepalli2021saint}), T2G-Former denoted as ``T2G'' (the attention-based model from \citet{yan2023t2g}), ExcelFormer denoted as ``Excel'' (the attention-based model from \citet{chen2023excelformer}), TabR (the retrieval-based model from \citet{gorishniy2023tabr}), ModernNCA denoted as ``MNCA'' (the retrieval-based model from \citet{ye2024modern}) and
GBDT, including XGBoost \citep{chen2016xgboost}, LightGBM \citep{ke2017lightgbm} and CatBoost \citep{prokhorenkova2018catboost}.

The models with non-linear feature embeddings from \citet{gorishniy2022embeddings} are marked with $\dagger$ or $\ddagger$ depending on the embedding type (see \autoref{A:sec:impl-feature-embeddings} for details on feature embeddings):
\begin{itemize}[nosep,leftmargin=2em]
    \item
    $\text{MLP}^\dagger$ and \modelminiemb\ use a modified version of the piecewise-linear embeddings.

    \item
    $\text{TabR}^\ddagger$, $\text{MNCA}^\ddagger$, and $\text{MLP}^\ddagger$ (also known as MLP-PLR) use various periodic embeddings.
\end{itemize}
More baselines are evaluated in \autoref{A:sec:extended-results}.
Implementation details are provided in \autoref{A:sec:implementation-details}.

\subsection{Task performance}
\label{sec:evaluation-performance}

\begin{figure}[h!]
    \centering
    \includegraphics[width=0.99\linewidth]{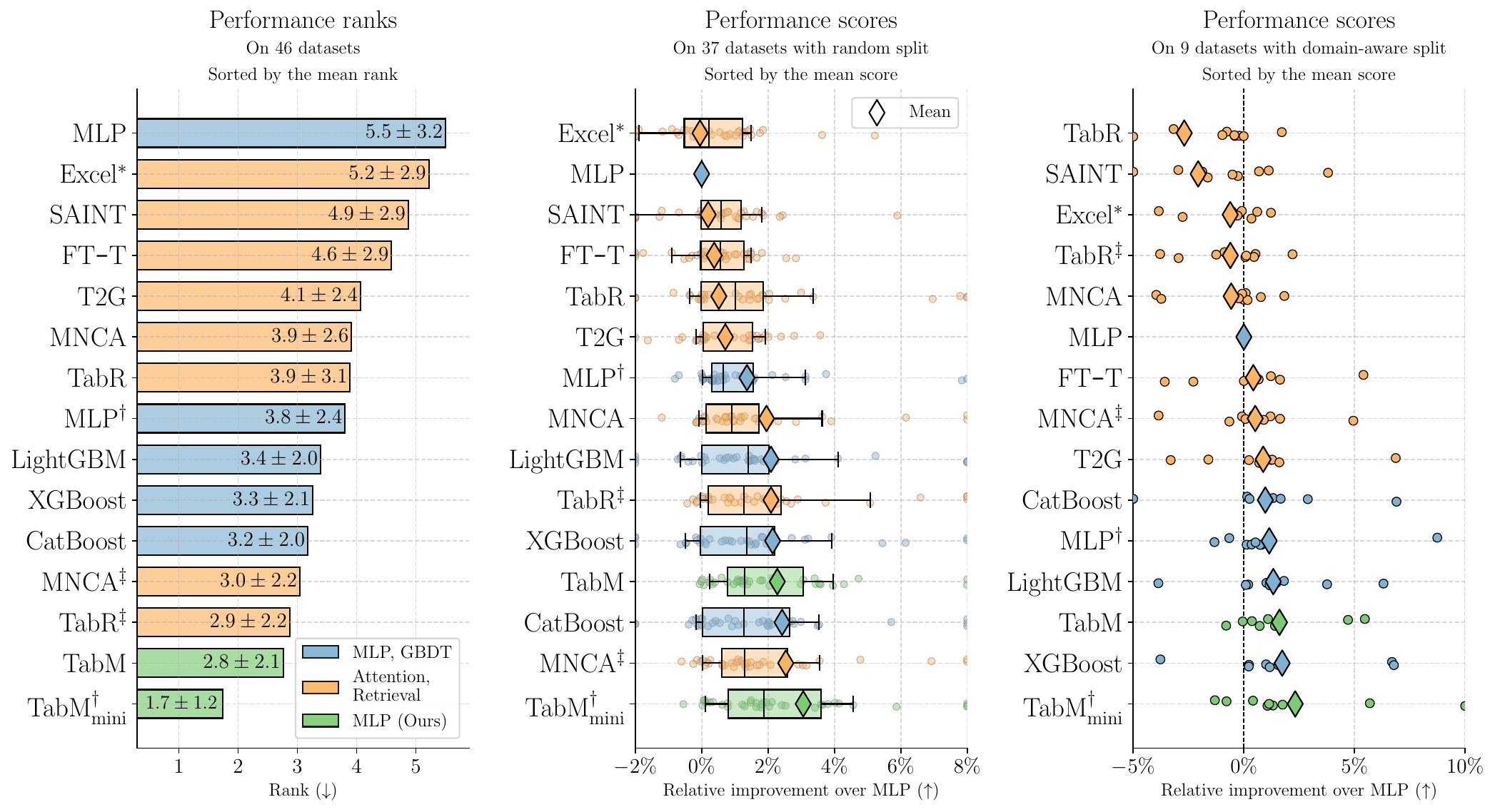}
    \caption{
        The task performance of tabular models on the \ndatasets\ datasets from \autoref{tab:datasets}.
        \textit{(Left)}
        The mean and standard deviations of the performance ranks over all datasets summarize the head-to-head comparison between the models on all datasets.
        \textit{(Middle \& Right)} The relative performance w.r.t. the plain multilayer perceptron (MLP) allows reasoning about the scale and consistency of improvements over this simple baseline.
        One dot of a jitter plot corresponds to the performance of a model on one of the \ndatasets\ datasets.
        The box plots visualize the 10th, 25th, 50th, 75th, and 90th percentiles of the jitter plots.
        Outliers are clipped.
        The separation in random and domain-aware dataset splits is explained in \autoref{sec:model-preliminaries}.
        (\mbox{$^*$Evaluated} under the common protocol without data augmentations)
    }
    \label{fig:performance}
\end{figure}

We evaluate all models following the protocol announced in \autoref{sec:model-preliminaries} and report the results in \autoref{fig:performance} (see also the critical difference diagram in \autoref{A:fig:cdd}).
We make the following observations:
\begin{enumerate}[nosep,leftmargin=2em]
    \item
    The performance ranks render \model\ as the top-tier DL model.

    \item
    The middle and right parts of \autoref{fig:performance} provide a fresh perspective on the per-dataset metrics.
    \model\ holds its leadership among the DL models.
    Meanwhile, many DL methods turn out to be no better or even worse than MLP on a non-negligible number of datasets, which shows them as less reliable solutions, and changes the ranking, especially on the domain-aware splits (right).

    \item
    One important characteristic of a model is the \textit{weakest} part of its performance profile (e.g. the 10th or 25th percentiles in the middle plot) since it shows how reliable the model is on ``inconvenient'' datasets.
    From that perspective, MLP\textsuperscript{\textdagger} seems to be a decent practical option between the plain MLP and \model, especially given its simplicity and efficiency compared to retrieval-based alternatives, such as TabR and ModernNCA.

\end{enumerate}

\textbf{Summary.}
\model\ confidently demonstrates the best performance among tabular DL models, and can serve as a reliable go-to DL baseline.
This is not the case for attention- and retrieval-based models.
Overall, MLP-like models, including \model, form a representative set of tabular DL baselines.

\subsection{Efficiency}
\label{sec:evaluation-efficiency}

Now, we evaluate tabular models in terms of training and inference efficiency, which becomes a serious reality check for some of the methods.
We benchmark exactly those hyperparameter configurations of models that are presented in \autoref{fig:performance} (see \autoref{a:sec:extended-efficiency} for the motivation).

\textbf{\modelminiembopt\ \& \modelminiembspadeopt.}
Additionally, in this section, we mark with the asterisk ($^*$) the versions of \model\ enhanced with two efficiency-related plugins available out-of-the-box in PyTorch \citep{paszke2019pytorch}: the automatic mixed precision (AMP) and \texttt{torch.compile} \citep{ansel2024pytorch2}.
The purpose of those \model\ variants is to showcase the potential of the modern hardware and software for a powerful tabular DL model, and they should not be directly compared to other DL models.
However, the implementation simplicity of \model\ plays an important role, because it facilitates the seamless integration of the aforementioned PyTorch plugins.

\begin{figure}[h!]
    \centering
    \begin{minipage}{0.47\textwidth}
        \centering
        \includegraphics[width=0.95\linewidth]{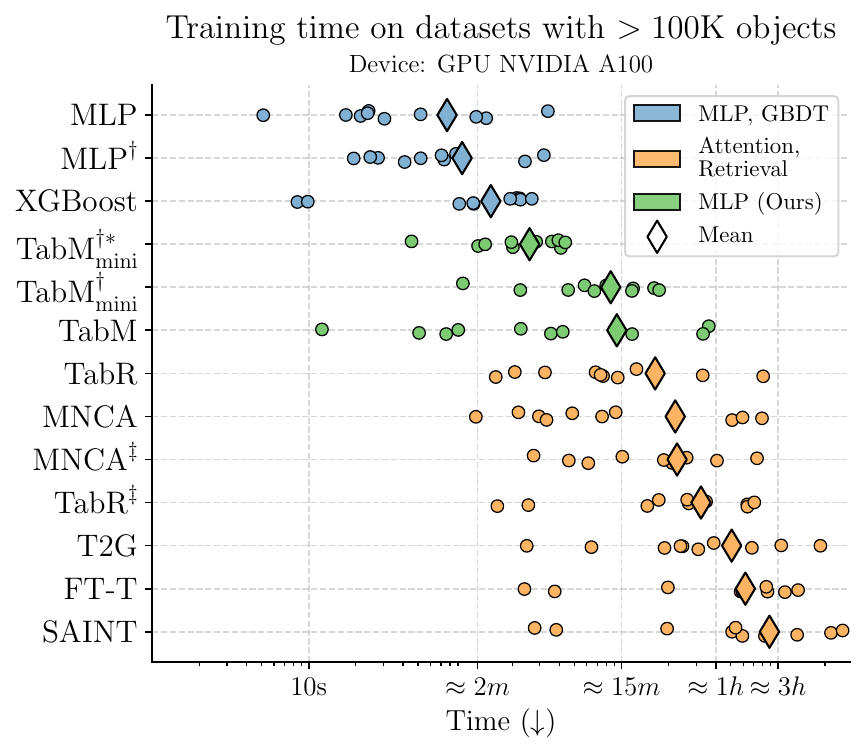}
    \end{minipage}
    \hfill
    \begin{minipage}{0.47\textwidth}
        \centering
        \includegraphics[width=0.95\linewidth]{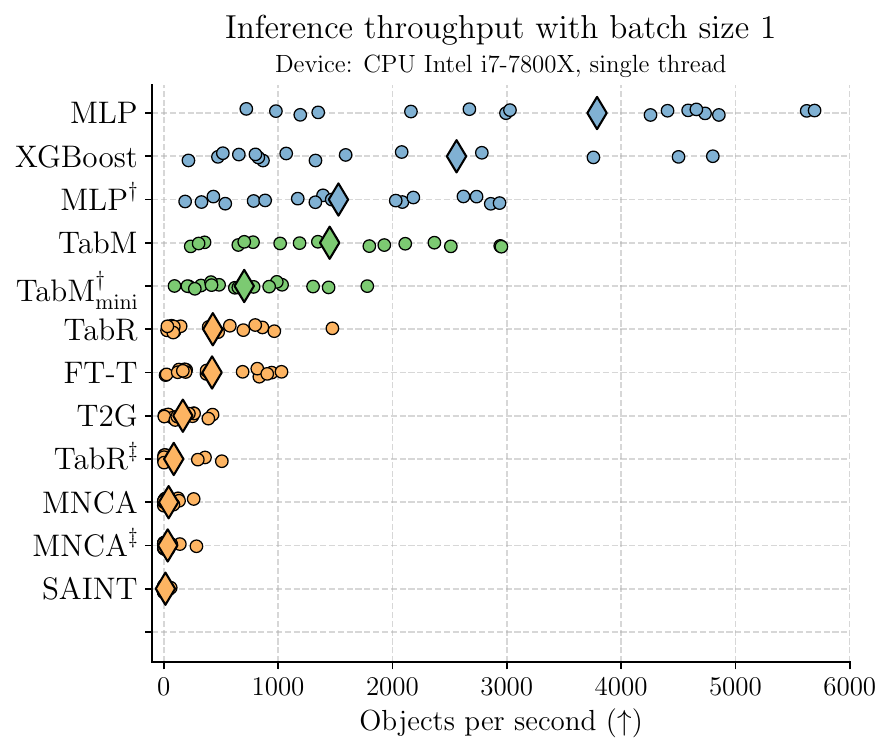}
    \end{minipage}
    \caption{
        Training times (left) and inference throughput (right) of the models from \autoref{fig:performance}.
        One dot represents a measurement on one dataset.
        \modelminiembopt\ is the optimized \modelminiemb\ (see \autoref{sec:evaluation-efficiency}).
    }
    \label{fig:efficiency}
\end{figure}

\textbf{Training time.}
We focus on training times on larger datasets, because on small datasets, all methods become almost equally affordable, regardless of the formal relative difference.
Nevertheless, in \autoref{A:fig:efficiency}, we provide measurements on small datasets as well.
The left side of \autoref{fig:efficiency} reveals that \model\ offers practical training times.
By contrast, the long training times of attention- and retrieval-based models become one more limitation of these methods.

\textbf{Inference throughput.}
The right side of \autoref{fig:efficiency} tells essentially the same story as the left side.
In \autoref{a:sec:extended-efficiency}, we also report the inference throughput on GPU with large batch sizes.

\textbf{Applicability to large datasets.}
In \autoref{tab:large}, we report metrics on two large datasets.
As expected, attention- and retrieval-based models struggle, yielding extremely long training times, or being simply inapplicable without additional effort.
See \autoref{A:sec:impl-evaluation-efficiency} for implementation details.

\textbf{Parameter count.}
Most tabular networks are overall compact.
This, in particular, applies to \model, because its size is by design comparable to MLP.
We report model sizes in \autoref{a:sec:extended-efficiency}.

\textbf{Summary.}
Simple MLPs are the fastest DL models, with \model\ being the runner-up.
The attention- and retrieval-based models are significantly slower.
Overall, MLP-like models, including \model, form a representative set of practical and accessible tabular DL baselines.

\begin{table}[h!]
\centering
\caption{
    RMSE (upper rows) and training times (lower rows) on two large datasets.
    The best values are in bold.
    The meaning of model colors follows \autoref{fig:performance}.
}
\label{tab:large}

\scalebox{0.875}{
\begin{tabular}{lcc|cccccc}
\toprule




& \#Objects
& \#Features
& {\transparent{1.0}\cellcolor[HTML]{C2DAEA}} {\color{black} $\mathrm{XGBoost}$}
& {\transparent{1.0}\cellcolor[HTML]{C2DAEA}} {\color{black} $\mathrm{MLP}$}
& {\transparent{1.0}\cellcolor[HTML]{C1E6BC}} {\color{black} \modelminiembspadeopt}
& {\transparent{1.0}\cellcolor[HTML]{C1E6BC}} {\color{black} \modelminiemb}
& {\transparent{1.0}\cellcolor[HTML]{FEDAB3}} {\color{black} $\mathrm{FT}\text{-}\mathrm{T}$}
& {\transparent{1.0}\cellcolor[HTML]{FEDAB3}} {\color{black} $\mathrm{TabR}$}
\\
\midrule
\multirow{ 2}{*}{Maps Routing}
& \multirow{ 2}{*}{$6.5$M}
& \multirow{ 2}{*}{$986$}
& $0.1601$
& $0.1592$
& $0.1583$
& $\mathbf{0.1582}$
& $0.1594$
& \multirow{ 2}{*}{OOM}
\\
&
&
& $28$m
& $\mathbf{15}$\textbf{m}
& $2$h
& $13.5$h
& $45.5$h
&
\\
\hline
\multirow{2}{*}{Weather}
& \multirow{2}{*}{$13$M}
& \multirow{2}{*}{$103$}
& $1.4234$
& $1.4842$
& $\mathbf{1.4090}$
& $\mathbf{1.4112}$
& $1.4409$
& \multirow{ 2}{*}{OOM}
\\
&
&
& $\mathbf{10}$\textbf{m}
& $15$m
& $1.3$h
& $3.3$h
& $13.5$h
&
\\
\bottomrule
\end{tabular}
}
\end{table}

\section{Analysis}
\label{sec:analysis}

\subsection{Performance and training dynamics of the individual submodels}
\label{sec:analysis-optimization}

Recall that the prediction of \model\ is defined as the mean prediction of its $k$ implicit submodels that share most of their weights.
In this section, we take a closer look at these submodels.

For the next experiment, we intentionally simplify the setup as described in detail in \autoref{A:sec:impl-analysis-optimization}.
Most importantly, all models have the same depth $3$ and width $512$, and are trained without early stopping, i.e. the training goes beyond the optimal epochs.
We use \modelmini\ from \autoref{fig:model} with $k=32$ denoted as \modelminik{32}.
We use \modelminik{1} (i.e. essentially one plain MLP) as a natural baseline for the submodels of \modelminik{32}, because each of the $32$ submodels has the architecture of \modelminik{1}.

\begin{figure*}[!h]
    \centering
    \includegraphics[width=0.99\linewidth]{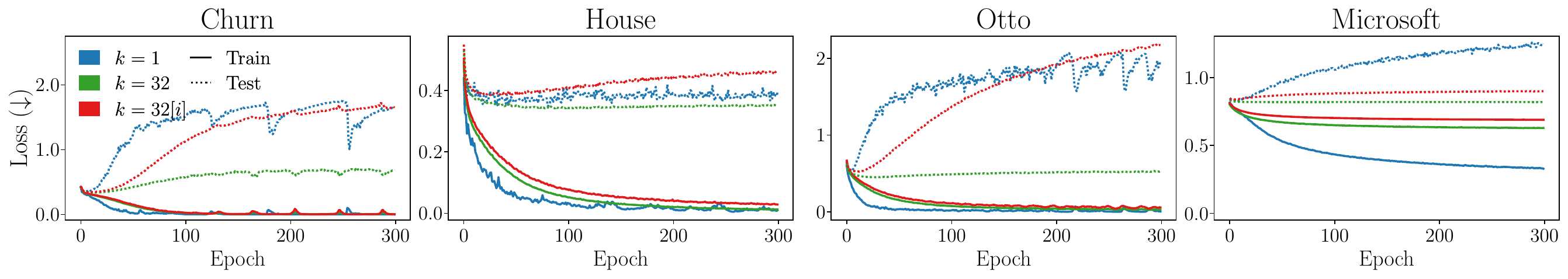}
    \includegraphics[width=0.99\linewidth]{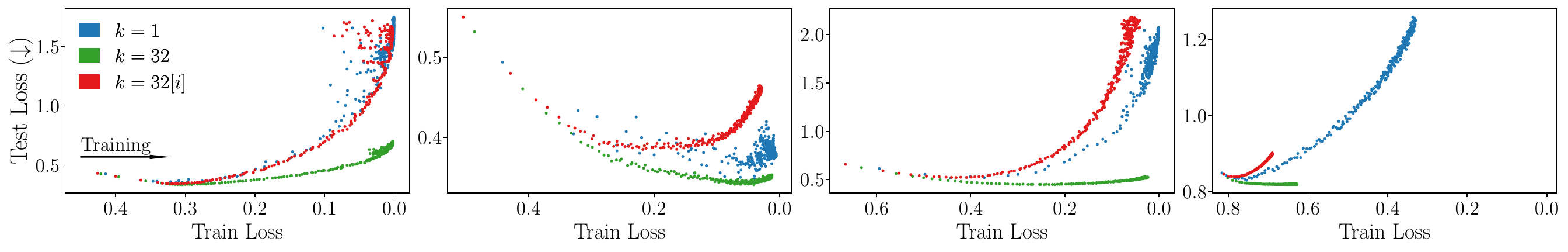}
    \caption{
        The training profiles of \modelminik{32} and \modelminik{1} as described in \autoref{sec:analysis-optimization}.
        \textit{(Upper)} The training curves. $k=32[i]$ represents the mean \textbf{i}ndividual loss over the $32$ submodels.
        \textit{(Lower)} Same as the first row, but in the train-test coordinates: each dot represents some epoch from the first row, and the training generally goes from left to right.
        This allows reasoning about overfitting by comparing test loss values for a given train loss value.
    }
    \label{fig:training-curves}
\end{figure*}

We visualize the training profiles on four diverse datasets (two classification and two regression problems of different sizes) in \autoref{fig:training-curves}.
As a reminder, the mean of the $k$ \textbf{\color{fontindividual} individual} losses is what is explicitly optimized during the training of \modelmini, the loss of the \textbf{\color{fontcollective} collective} mean prediction corresponds to how \modelmini\ makes predictions on inference, and \modelminik{1} is just a \textbf{\color{fontmlp} baseline}.

In the upper row of \autoref{fig:training-curves}, the collective mean prediction of the submodels is superior to their individual predictions in terms of both training and test losses.
After the initial epochs, the training loss of the baseline MLP is lower than that of the collective and individual predictions.

In the lower row of \autoref{fig:training-curves}, we see a stark contrast between the individual and collective performance of the submodels.
Compared to the baseline MLP, the submodels look overfitted individually, while their collective prediction exhibits substantially better generalization.
This result is strict evidence of a non-trivial diversity of the submodels: without that, their collective test performance would be similar to their individual test performance.
Additionally, we report the performance of the \textbf{B}est submodel of \model\
across many datasets under the name \modelbesthead\ in \autoref{fig:submodels}.
As such, individually, even the best submodel of \model\ is no better than a simple MLP.

\textbf{Summary.}
\model\ draws its power from the collective prediction of weak, but diverse submodels.

\subsection{Selecting submodels after training}
\label{sec:analysis-selecting-submodels}

The design of \model\ allows selecting only a subset of submodels after training based on any criteria, simply by pruning extra prediction heads and the corresponding rows of the adapter matrices.
To showcase this mechanics, after the training, we \textbf{G}reedily construct a subset of \model's submodels with the best collective performance on the validation set, and denote this ``pruned'' \model\ as \modelgreedyheads.
The performance reported in \autoref{fig:submodels} shows that \modelgreedyheads\ is slightly behind the vanilla \model.
On average over \ndatasets\ datasets, the greedy submodel selection results in $8.8 \pm 6.6$ submodels out of the initial $k=32$, which can result in faster inference.
See \autoref{A:sec:impl-analysis-selecting-submodels} for implementation details.

\begin{minipage}{0.49\linewidth}
    \centering
    \includegraphics[width=0.95\linewidth]{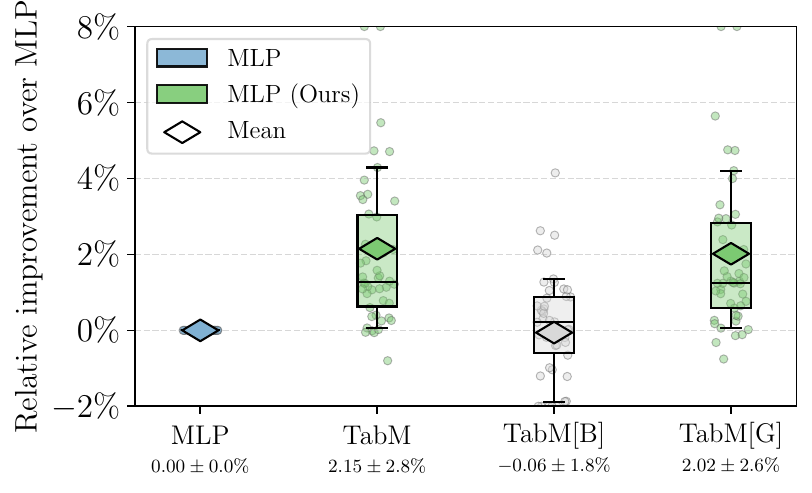}
    \captionof{figure}{
        The performance on the \ndatasets\ datasets from \autoref{tab:datasets}.
        \modelbesthead\ and \modelgreedyheads\ are described in \autoref{sec:analysis-optimization} and \autoref{sec:analysis-selecting-submodels}.
    }
    \label{fig:submodels}
\end{minipage}
\hfill
\begin{minipage}{0.45\linewidth}
    \centering
    \includegraphics[width=0.97\linewidth]{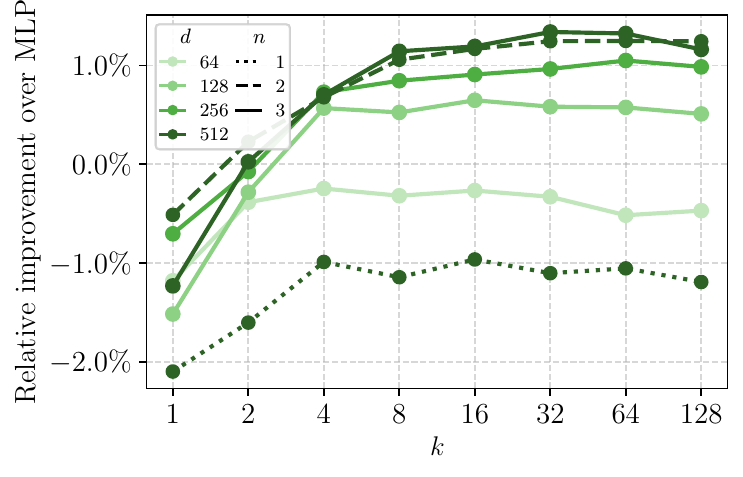}
    \captionof{figure}{
        The average performance of \model\ with $n$ layers of the width $d$ across $17$ datasets as a function of $k$.
    }
    \label{fig:d-k-ablation}
\end{minipage}

\subsection{How does the performance of \model\ depend on $k$?}
\label{sec:analysis-k}

To answer the question in the title, we consider \model\ with $n$ layers of the size $d$ and different values of $k$, and report the average performance over multiple datasets in \autoref{fig:d-k-ablation} (the implementation details are provided in \autoref{A:sec:impl-analysis-k}).
The solid curves correspond to $n = 3$, and the dark green curves correspond to $d = 512$.
Our main observations are as follows.
\textit{First,} it seems that the ``larger'' \model\ is (i.e. when $n$ and $d$ increase), the more submodels it can accommodate effectively.
For example, note how the solid curves corresponding to different $d$ diverge at $k = 2$ and $k = 4$.
\textit{Second,} too high values of $k$ can be detrimental.
Perhaps, weight sharing limits the number of submodels that can productively ``coexist'' in one network, despite the presence of non-shared adapters.
\textit{Third}, too narrow ($d = 64$) or too shallow ($n = 1$) configurations of TabM can lead to suboptimal performance, at least in the scope of middle-to-large datasets considered in this work.

\subsection{Parameter-efficient ensembling reduces the number of dead neurons}
\label{sec:analysis-dead-neurons}

Here, we show empirically that the design of \model\ naturally leads to higher utilization of the backbone's weights.
Even without technical definitions, this sounds intuitive, since \model\ has to implement $k$ (diverse) computations using the amount of weights close to that of one MLP.

Let's consider \modelmini\ as illustrated in \autoref{fig:model}.
By design, each of the shared neurons of \modelmini\ is used $k$ times per forward pass, where ``neuron'' refers to the combination of the linear transformation and the subsequent nonlinearity (e.g. ReLU).
By contrast, in plain MLP (or in \modelmini\ with $k=1$), each neuron is used only once per forward pass.
Thus, technically, a neuron in \modelmini\ has more chances to be activated, which overall may lead to lower portion of dead neurons in \modelmini\ compared to MLP (a dead neuron is a neuron that never activates, and thus has no impact on the prediction).
Using the experiment setup from \autoref{sec:analysis-optimization}, we compute the portion of dead neurons in \modelmini\ using its best validation checkpoint.
On average across \ndatasets\ datasets, for $k = 1$ and $k = 32$, we get $0.29 \pm 0.17$ and $0.14 \pm 0.09$ portion of dead neurons, respectively, which is in line with the described intuition.
Technically, on a given dataset, this metric is computed as the percentage of neurons that never activate on a fixed set of $2048$ training objects.

\section{Conclusion \& Future work}
\label{sec:conclusion}

In this work, we have demonstrated that tabular multilayer perceptrons (MLPs) greatly benefit from parameter-efficient ensembling.
Using this insight, we have developed \model\ --- a simple MLP-based model with state-of-the-art performance.
In a large-scale comparison with many tabular DL models, we have demonstrated that \model\ is ready to serve as a new powerful and efficient tabular DL baseline.
Along the way, we highlighted the important technical details behind \model\ and discussed the individual performance of the implicit submodels underlying \model.

One idea for future work is to bring the power of (parameter-)efficient ensembles to other, non-tabular, domains with optimization-related challenges and, ideally, lightweight base models.
Another idea is to evaluate \model\ for uncertainty estimation and out-of-distribution (OOD) detection on tabular data, which is inspired by works like \citet{lakshminarayanan2017simple}.

\newpage
\textbf{Reproducibility statement.}
The code is provided in the following repository: \href{\repository}{link}.
It contains the implementation of \model, hyperparameter tuning scripts, evaluation scripts, configuration files with hyperparameters (the TOML files in the \texttt{exp/} directory), and the report files with the main metrics (the JSON files in the \texttt{exp/} directory).
In the paper, the model is described in \autoref{sec:model}, and the implementation details are provided in \autoref{A:sec:implementation-details}.

\bibliography{references}
\bibliographystyle{iclr2025_conference}
\newpage

\appendix

\section{Additional discussion on \model}

\subsection{Motivation}
\label{A:sec:model-motivation}

\textbf{Why BatchEnsemble?}
Among relatively ease-to-use ``efficient ensembling'' methods, beyond BatchEnsemble, there are examples such as dropout ensembles \citep{lakshminarayanan2017simple}, naive multi-head architectures, TreeNet \citep{lee2015why}.
However, in the literature, they were consistently outperformed by more advanced methods, including BatchEnsemble \citep{wen2020batchensemble}, MIMO \citep{havasi2020training}, FiLM-Ensemble \citep{turkoglu2022film}.

Among advanced methods, BatchEnsemble seems to be one of the simplest and most flexible options.
For example, FiLM-Ensemble \citep{turkoglu2022film} requires normalization layers to be presented in the original architecture, which is not always the case for tabular MLPs.
MIMO \citep{havasi2020training}, in turn, imposes additional limitations compared to BatchEnsemble.
\textit{First}, it requires \textit{concatenating} (not \textit{stacking}, as with BatchEnsemble) all $k$ input representations, which increases the input size of the first linear layer.
With the relatively high number of submodels $k = 32$ used in our paper, this can be an issue on datasets with a large number of features, especially when feature embeddings \citep{gorishniy2022embeddings} are used.
For example, for $k = 32$, the number of features $m = 1000$, and the feature embedding size $l = 32$, the input size approaches one million resulting in an extremely large first linear layer of MLP.
\textit{Second}, with BatchEnsemble, it is easy to explicitly materialize, analyze, and prune individual submodels.
By contrast, in MIMO, all submodels are implicitly entangled within one MLP, and there is no easy way to access individual submodels.

\textbf{Why MLPs?}
Despite the applicability of BatchEnsemble \citep{wen2020batchensemble} to almost any architecture, we focus specifically on MLPs.
The key reason is \textit{efficiency}.
\textit{First,} to achieve high performance, throughout the paper, we use the relatively large number of submodels $k = 32$.
However, the desired less-than-$\times k$ runtime overhead of BatchEnsemble typically happens only when the original model underutilizes the power of parallel computations of a given hardware.
This will not be the case for attention-based models on datasets with a large number of features, as well as for retrieval-based models on datasets with a large number of objects.
\textit{Second,} as we show in \autoref{sec:evaluation-efficiency}, attention- and retrieval-based models are already slow as-is.
By contrast, MLPs are exceptionally efficient, to the extent that slowing them down even by an order of magnitude will still result in practical models.

Also, generally speaking, the definition of MLP suggested in \autoref{sec:model-design} and used in \model\ is not special, and more advanced MLP-like backbones can be used.
However, in preliminary experiments, we did not observe the benefits of more advanced backbones.
Perhaps, small technical differences between backbones become less impactful in the context of parameter-efficient ensembling, at least in the scope of middle-to-large-sized datasets.

\subsection{\model\ with feature embeddings}
\label{A:sec:model-emb}

\textbf{Notation.}
In this paper, we use $\dagger$ to mark \model\ variants with the piecewise-linear embeddings (e.g. \modelminiemb, \modelemb, etc.).

\textbf{Implementation details.}
In fact, there are no changes in the usage of feature embeddings compared to plain MLPs:
feature embeddings are applied, and the result is flattened, before being passed to the backbones in terms of \autoref{fig:model}.
For example, if a dataset has $m$ continuous features and all of them are embedded, the very first adapter $R$ will have the shape $k \times md_e$, where $d_e$ is the feature embedding size.
For \modelminiemb\ and \modelemb, we initialize the first multiplicative adapter $R$ of the first linear layer from the standard normal distribution $\mathcal{N}(0, 1)$.
The remaining details are best understood from the source code.

\textbf{Efficiency.}
When feature embeddings are used, the simplified batching strategy from \autoref{sec:model-important} allows for more efficient implementation, when the feature embeddings are applied to the original \texttt{batch\_size} objects, and the result is simply cloned $k$ times (compared to embedding $k \times \texttt{batch\_size}$ objects with the original batching strategy).

\subsection{Hyperparameters}
\label{A:sec:model-hyperparameters}

We noticed that the typical optimal learning rate for \model\ is higher than for MLP (note that, on each dataset, the batch size is the same for all DL models).
We hypothesize that the reason is the effectively larger batch size for \model\ because of how the training batches are constructed (even if the simplified batching strategy from \autoref{sec:model-important} is used).

\subsection{Limitations and practical considerations}
\label{A:sec:model-limitations}

\model\ does not introduce any new limitations compared to BatchEnsemble \citep{wen2020batchensemble}.
Nevertheless, we note the following:
\begin{itemize}[nosep,leftmargin=2em]
    \item
    The MLP backbone used in \model\ is one of the simplest possible, and generally, more advanced backbones can be used.
    That said, some backbones may require additional care when used in \model.
    For example, we did not explore backbones with normalization layers.
    For such layers, it is possible to allocate non-shared trainable affine transformations for each implicit submodel by adding one multiplicative and one additive adapter after the normalization layer (i.e. like in FiLM-Ensemble \citep{turkoglu2022film}).
    Additional experiments are required to find the best strategy.

    \item
    For ensemble-like models, such as \model, the notion of ``the final object embedding`` changes: now, it is not a single vector, but a set of $k$ vectors.
    If exactly one object embedding is required, then additional experiments may be needed to find the best way to combine $k$ embeddings into one.
    The presence of multiple object embeddings can also be important for scenarios when \model\ is used for solving more than one task, in particular when it is pretrained as a generic feature extractor and then reused for other tasks.
    The main practical guideline is that the $k$ prediction branches should not interact with each other (e.g. through attention, pooling, etc.) and should always be trained separately.
\end{itemize}

\section{Extended results}
\label{A:sec:extended-results}

This section complements \autoref{sec:evaluation}.

\subsection{Additional baselines}

In addition to the models from \autoref{sec:evaluation-baselines}, we consider the following baselines:

\begin{itemize}[nosep,leftmargin=2em]
    \item MLP-PLR \cite{gorishniy2022embeddings}, that is, an MLP with periodic embeddings.
    \item ResNet \citep{gorishniy2021revisiting}
    \item SNN \citep{klambauer2017self}
    \item DCNv2 \citep{wang2020dcn2}
    \item AutoInt \citep{song2019autoint}
    \item MLP-Mixer is our adaptation of \cite{tolstikhin2021mlp} for tabular data.
    \item Trompt \citep{chen2023trompt} (our reimplementation, since there is no official implementation)
\end{itemize}

We also evaluated TabPFN \citep{hollmann2022tabpfn}, where possible.
The results for this model are available only in \autoref{A:sec:per-dataset-results} because this model is by design not applicable to regression tasks, which is a considerable number of our datasets.
Overall, TabPFN specializes in small datasets.
In line with that, the performance of TabPFN on our benchmark was not competitive.

\subsection{Task performance}

\begin{figure}[h!]
    \centering
    \includegraphics[width=0.97\linewidth]{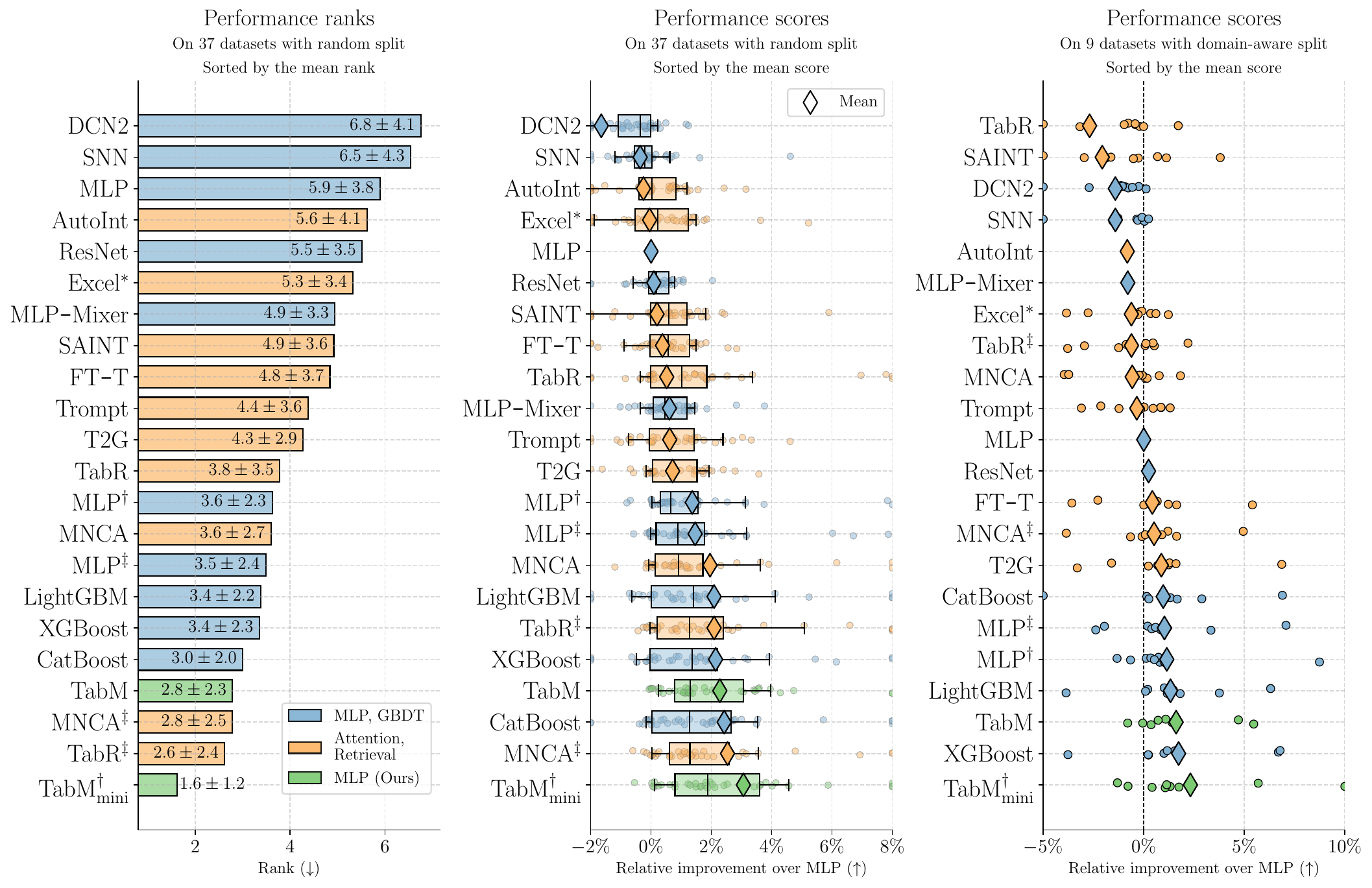}
    \caption{
        An extended comparison of tabular models as in \autoref{fig:performance}.
        Note that the ranks (left) are computed only over the \nrandomsplits\ datasets with random splits because ResNet, AutoInt, and MLP-Mixer were evaluated only on one $1$ out of $9$ datasets with domain-aware splits.
    }
    \label{A:fig:main-comparison}
\end{figure}

\autoref{A:fig:main-comparison} is a different version of \autoref{fig:performance} with additional baselines.
Overall, none of the additional baselines affect our main story.

\begin{figure}[h!]
    \centering
    \includegraphics[width=0.6\linewidth]{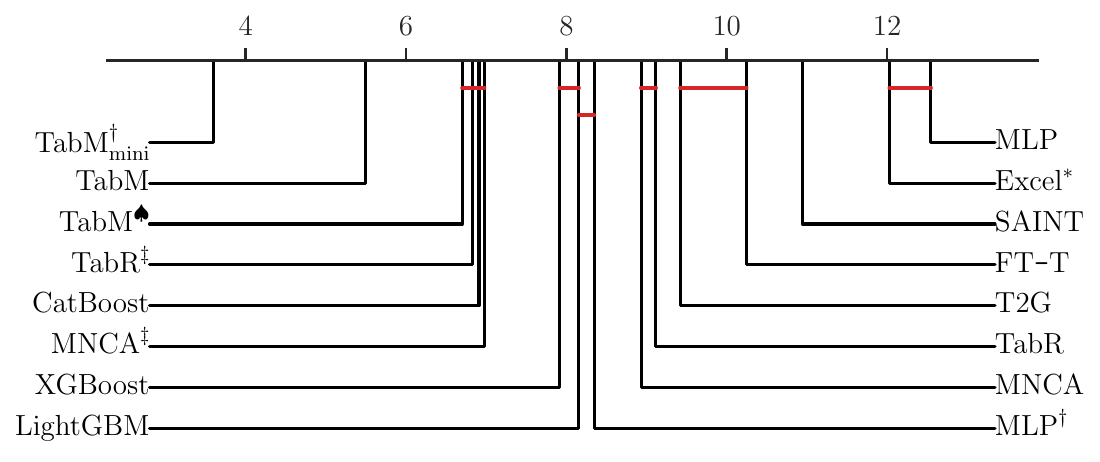}
    \caption{
        Critical difference diagram.
        The computation method is taken from the \cite{kim2024carte}.
    }
    \label{A:fig:cdd}
\end{figure}

\autoref{A:fig:cdd} is the critical difference diagram (CDD) computed over exactly the same results that were used for building \autoref{fig:performance}.

\newpage
\subsection{Efficiency}
\label{a:sec:extended-efficiency}

This section complements \autoref{sec:evaluation-efficiency}.

\textbf{Additional results.}

\begin{figure}[h!]
    \centering
    \begin{minipage}{0.47\textwidth}
        \centering
        \includegraphics[width=0.95\linewidth]{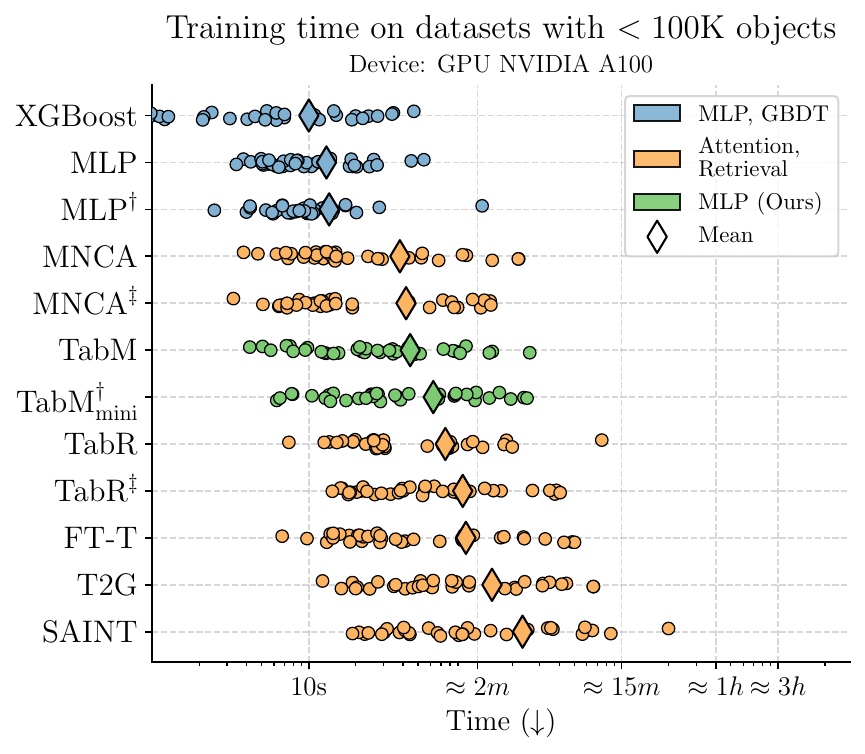}
    \end{minipage}
    \hfill
    \begin{minipage}{0.47\textwidth}
        \centering
        \includegraphics[width=0.95\linewidth]{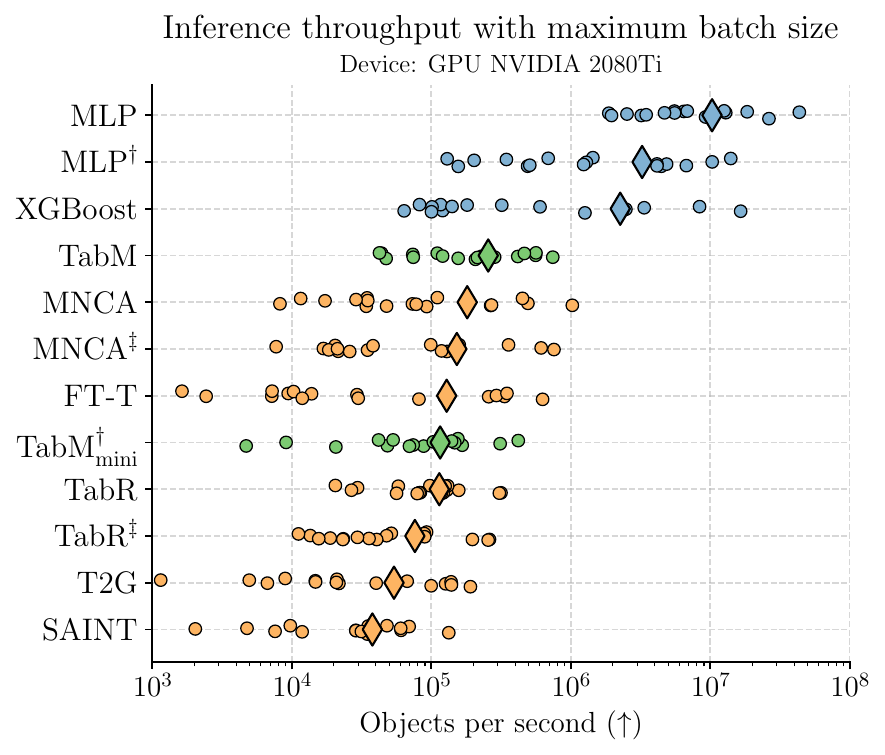}
    \end{minipage}
    \caption{
        (\textit{Left}) Training time on datasets with less than 100K objects.
        (\textit{Right}) Inference throughput on GPU with maximum possible batch size (i.e. the batch size depends on a model).
    }
    \label{A:fig:efficiency}
\end{figure}

\autoref{A:fig:efficiency} complements \autoref{fig:efficiency} by providing the training times on smaller datasets and the inference throughput on GPU with large batch sizes.

\autoref{A:tab:n-params} provide the number of trainable parameters for some of the models from \autoref{fig:performance}.

\textbf{Motivation for the benchmark setup.}
Comparing models under all possible kinds of budgets (task performance, the number of parameters, training time, etc.) on all possible hardware (GPU, CPU, etc.) with all possible batch sizes is rather infeasible.
As such, we set a narrow goal of \textit{providing a high-level intuition on the efficiency in a transparent setting}.
Thus, benchmarking the transparently obtained tuned hyperparameter configurations works well for our goal.
Yet, this choice also has a limitation: the hyperparameter tuning process is not aware of the efficiency budget, so it can prefer much heavier configurations even if they lead to tiny performance improvements, which will negatively affect efficiency without a good reason.
Overall, we hope that the large number of datasets compensates for potentially imperfect per-dataset measurements.

\textbf{Motivation for the two setups for measuring inference throughput.}
\begin{itemize}[nosep,leftmargin=2em]
    \item The setup on the right side of \autoref{fig:efficiency} simulates the online per-object predictions.
    \item The setup on the right side of \autoref{A:fig:efficiency} simulates the offline batched computations.
\end{itemize}

\begin{table}[h!]
    \centering
    \caption{Mean number of parameters with std. dev. for 7 different tuned models across all \ndatasets\ datasets.}
    \scalebox{0.75}{\begin{tabular}{ccccccc}
\toprule
\model & MLP & FT-T & T2G & TabR & ModernNCA & SAINT \\
\midrule
$1.4M\pm1.3M$ &
$1.0M\pm1.0M$ &
$1.2M\pm1.2M$ &
$2.1M\pm1.6M$ &
$858K\pm1.4M$ &
$1.0M\pm1.1M$ &
$175.4M\pm565.4M$ \\
\bottomrule
    
\end{tabular}}
    \label{A:tab:n-params}
\end{table}

\section{Datasets}
\label{A:sec:datasets}

In total, we use \ndatasets\ datasets:
\begin{enumerate}[nosep,leftmargin=2em]
    \item
    $38$ datasets are taken from \cite{gorishniy2023tabr}, which includes:

    \begin{enumerate}[nosep,leftmargin=2em]
        \item
        $28$ datasets from \cite{grinsztajn2022why}.
        See the original paper for the precise dataset information.

        \item
        $10$ datasets from other sources.
        Their properties are provided in \autoref{A:tab:default-datasets}.

    \end{enumerate}

    \item
    $8$ datasets from the TabReD benchmark \citep{rubachev2024tabred}.
    Their properties are provided in \autoref{A:tab:tabred-datasets}.
\end{enumerate}

In fact, the aforementioned $38$ datasets from \citet{gorishniy2023tabr} is only a subset of the datasets used in \citet{gorishniy2023tabr}.
Namely, we did not include the following of the remaining datasets:

\begin{itemize}[nosep,leftmargin=2em]
    \item
    The datasets that, according to \citet{rubachev2024tabred}, have incorrect splits and/or label leakage, including:
    $\mathrm{Bike\_Sharing\_Demand}$,
    $\mathrm{compass}$,
    $\mathrm{electricity}$,
    $\mathrm{SGEMM\_GPU\_kernel\_performance}$,
    $\mathrm{sulfur}$,
    $\mathrm{visualizing\_soil}$,
    and the weather forecasting dataset (it is replaced by the correct weather forecasting dataset from TabReD \citep{rubachev2024tabred}).

    \item $\mathrm{rl}$ from \citep{grinsztajn2022why}.
    We observed abnormal results on these datasets.
    This is an anonymous dataset, which made the investigation impossible, so we removed this dataset to avoid confusion.

    \item
    $\mathrm{yprop\_4\_1}$ from \citep{grinsztajn2022why}.
    Strictly speaking, this dataset was omitted due to a mistake on our side.
    For future work, we note that the typical performance gaps on this dataset have low absolute values in terms of RMSE.
    Perhaps, $R^2$ may be a more appropriate metric for this dataset.

\end{itemize}

\begin{table*}[h!]
    \setlength\tabcolsep{2.2pt}
    \centering
    \caption{
        Properties of those datasets from \citet{gorishniy2023tabr}
        that are not part of \citet{grinsztajn2022why} or TabReD \citet{rubachev2024tabred}.
        ``\# Num'', ``\# Bin'', and ``\# Cat'' denote the number of numerical, binary, and categorical features, respectively.
        The table is taken from \citep{gorishniy2023tabr}.
    }
    \label{A:tab:default-datasets}
    \scalebox{0.9}{\begin{tabular}{llcccccclc}
\toprule
Name & \# Train & \# Validation & \# Test & \# Num & \# Bin & \# Cat & Task type & Batch size \\
\midrule
Churn Modelling & $6\,400$ & $1\,600$ & $2\,000$ & $7$ & $3$ & $1$ & Binclass & 128 \\
California Housing & $13\,209$ & $3\,303$ & $4\,128$ & $8$ & $0$ & $0$ & Regression & 256 \\
House 16H & $14\,581$ & $3\,646$ & $4\,557$ & $16$ & $0$ & $0$ & Regression & 256 \\
Adult & $26\,048$ & $6\,513$ & $16\,281$ & $6$ & $1$ & $8$ & Binclass & 256 \\
Diamond & $34\,521$ & $8\,631$ & $10\,788$ & $6$ & $0$ & $3$ & Regression & 512 \\
Otto Group Products & $39\,601$ & $9\,901$ & $12\,376$ & $93$ & $0$ & $0$ & Multiclass & 512 \\
Higgs Small & $62\,751$ & $15\,688$ & $19\,610$ & $28$ & $0$ & $0$ & Binclass & 512 \\
Black Friday & $106\,764$ & $26\,692$ & $33\,365$ & $4$ & $1$ & $4$ & Regression & 512 \\
Covertype & $371\,847$ & $92\,962$ & $116\,203$ & $10$ & $4$ & $1$ & Multiclass & 1024 \\
Microsoft & $723\,412$ & $235\,259$ & $241\,521$ & $131$ & $5$ & $0$ & Regression & 1024 \\
\bottomrule
\end{tabular}

}
\end{table*}

\begin{table*}[h!]
    \setlength\tabcolsep{2.2pt}
    \centering
    \caption{
        Properties of the datasets from the TabReD benchmark \citep{rubachev2024tabred}.
        ``\# Num'', ``\# Bin'', and ``\# Cat'' denote the number of numerical, binary, and categorical features, respectively.
    }
    \label{A:tab:tabred-datasets}
    \scalebox{0.9}{\begin{tabular}{llcccccclc}
\toprule
Name & \# Train & \# Validation & \# Test & \# Num & \# Bin & \# Cat & Task type & Batch size \\
\midrule
Sberbank Housing & $18\,847$ & $4\,827$ & $4\,647$ & $365$ & $17$ & $10$ & Regression & 256 \\
Ecom Offers & $109\,341$ & $24\,261$ & $26\,455$ & $113$ & $6$ & $0$ & Binclass & 1024 \\
Maps Routing & $160\,019$ & $59\,975$ & $59\,951$ & $984$ & $0$ & $2$ & Regression & 1024 \\
Homesite Insurance & $224\,320$ & $20\,138$ & $16\,295$ & $253$ & $23$ & $23$ & Binclass & 1024 \\
Cooking Time & $227\,087$ & $51\,251$ & $41\,648$ & $186$ & $3$ & $3$ & Regression & 1024 \\
Homecredit Default & $267\,645$ & $58\,018$ & $56\,001$ & $612$ & $2$ & $82$ & Binclass & 1024 \\
Delivery ETA & $279\,415$ & $34\,174$ & $36\,927$ & $221$ & $1$ & $1$ & Regression & 1024 \\
Weather & $106\,764$ & $42\,359$ & $40\,840$ & $100$ & $3$ & $0$ & Regression & 1024 \\
\bottomrule
\end{tabular}

}
\end{table*}

\section{Implementation details}
\label{A:sec:implementation-details}

\subsection{Hardware}
\label{A:sec:impl-hardware}

Most of the experiments were conducted on a single NVIDIA A100 GPU.
In rare exceptions, we used a machine with a single NVIDIA 2080 Ti GPU and Intel(R) Core(TM) i7-7800X CPU @ 3.50GHz.

\subsection{Experiment setup}
\label{A:sec:impl-experiment-setup}

We mostly follow the experiment setup from \cite{gorishniy2023tabr}.
As such, some of the text below is copied from \citep{gorishniy2023tabr}.

\textbf{Data preprocessing.}
For each dataset, for all DL-based solutions, the same preprocessing was used for fair comparison.
For numerical features, by default, we used a slightly modified version of the quantile normalization from the Scikit-learn package \citep{pedregosa2011scikit} (see the source code), with rare exceptions when it turned out to be detrimental (for such datasets, we used the standard normalization or no normalization).
For categorical features, we used one-hot encoding.
Binary features (i.e. the ones that take only two distinct values) are mapped to $\{0,1\}$ without any further preprocessing. We completely follow \cite{rubachev2024tabred} on \autoref{A:tab:tabred-datasets} datasets.

\textbf{Training neural networks.}
For DL-based algorithms, we minimize cross-entropy for classification problems and mean squared error for regression problems.
We use the AdamW optimizer \citep{loshchilov2019decoupled}.
We do not apply learning rate schedules.
We do not use data augmentations.
We apply global gradient clipping to $1.0$.
For each dataset, we used a predefined dataset-specific batch size.
We continue training until there are $\texttt{patience}$ consecutive epochs without improvements on the validation set; we set $\texttt{patience} = 16$ for the DL models.

\textbf{Hyperparameter tuning.}
In most cases, hyperparameter tuning is performed with the TPE sampler (typically, 50-100 iterations) from the Optuna package \citep{akiba2019optuna}.
Hyperparameter tuning spaces for most models are provided in individual sections below (example for \model: \autoref{A:sec:impl-model}). We follow \cite{rubachev2024tabred} and use $25$ iterations on some datasets from \autoref{A:tab:tabred-datasets}.

\textbf{Evaluation.}
On a given dataset, for a given model, the tuned hyperparameters are evaluated under multiple (in most cases, $15$) random seeds.
The mean test metric and its standard deviation over these random seeds are then used to compare algorithms as described in \autoref{A:sec:impl-metrics}.

\subsection{Metrics}
\label{A:sec:impl-metrics}

\begin{figure}
    \centering
    \includegraphics[width=0.95\linewidth]{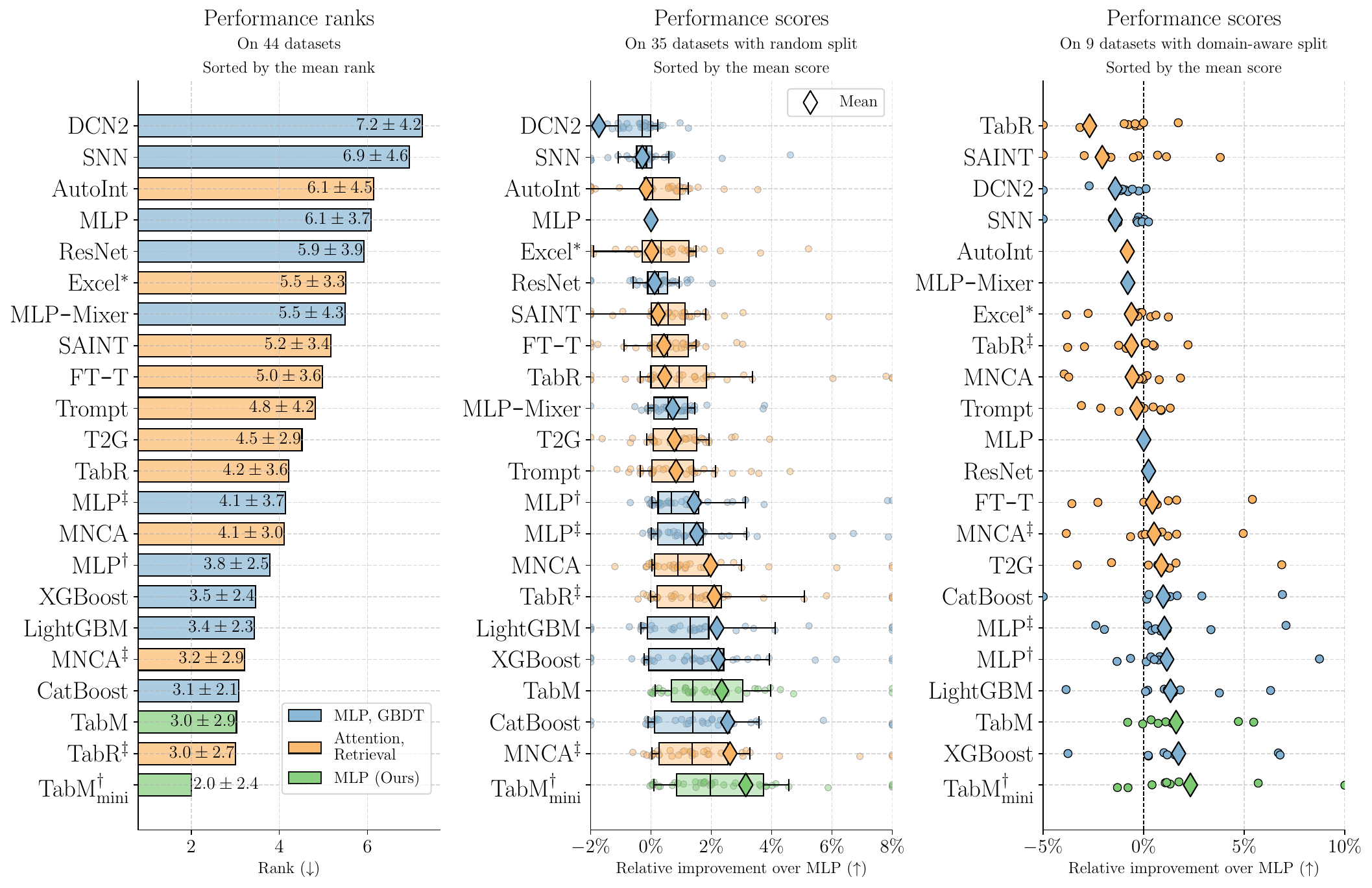}
    \caption{
        Same as \autoref{fig:performance}, but ROC-AUC is used as the metric for all classification datasets.
        The two multiclass datasets presented in our benchmark are not taken into account.
    }
    \label{A:fig:evaluation-roc-auc}
\end{figure}

We use Root Mean Squared Error for regression tasks, ROC-AUC for classification datasets from \autoref{A:tab:tabred-datasets} (following \citet{rubachev2024tabred}), and accuracy for the rest of datasets (following \citet{gorishniy2023tabr}).
We also tried computing ROC-AUC for all classification datasets, but did not observe any significant changes (see \autoref{A:fig:evaluation-roc-auc}), so we stuck to prior work.
By default, the mean test score and its standard deviation are obtained by training a given model with tuned hyperparameters from scratch on a given dataset under 15 different random seeds.

\textbf{How we compute ranks.}
Our method of computing ranks used in \autoref{fig:performance} does not count small improvements as wins, hence the reduced range of ranks compared to other studies.
Intuitively, our ranks can be considered as “tiers”.

Recall that, on a given dataset, the performance of a given model A is expressed with the mean $\text{A}_\text{mean}$ and the standard deviation $\text{A}_\text{std}$ of the performance score computed after the evaluation under multiple random seeds.
Assuming the higher score the better, we define that the model A is better than the model B if: $\text{A}_\text{mean} - \text{A}_\text{std} > \text{B}_\text{mean}$.
In other words, a model is considered better if it has a better mean score and the margin is larger than the standard deviation.

On a given dataset, when there are many models, we sort them in descending score order.
Starting from the best model (with a rank equal to $1$) we iterate over models and assign the rank $1$ to all models that are no worse than the best model according to the above rule.
The first model in descending order that is worse than the best model is assigned rank $2$ and becomes the new reference model.
We continue the process until all models are ranked.
Ranks are computed independently for each dataset.

\subsection{Implementation details of \autoref{sec:evaluation-efficiency}}
\label{A:sec:impl-evaluation-efficiency}

\textbf{Applicability to large datasets.}
The two datasets used in \autoref{tab:large} are the \textit{full} versions of the ``Weather'' and ``Maps Routing'' datasets from the TabReD benchmark \cite{rubachev2024tabred}.
Their smaller versions with subsampled training set were already included in \autoref{tab:datasets} and were used when building \autoref{fig:performance}.
The validation and test sets are the same for the small and large versions of these datasets, so the task metrics are comparable between the two versions.
When running models on the large versions of the datasets, we reused the hyperparameters tuned for their small versions.
Thus, this experiment can be seen as a quick assessment of the applicability of several tabular DL to large datasets without a strong focus on the task performance.
All models, except for FT-Transformer, were evaluated under $3$ random seeds.
FT-Transformer was evaluated under $1$ random seed.

\subsection{Implementation details of \autoref{sec:analysis-optimization}}
\label{A:sec:impl-analysis-optimization}

\textbf{Experiment setup.}
This paragraph complements the description of the experiment setup in \autoref{sec:analysis-optimization}.
Namely, in addition to what is mentioned in the main text:
\begin{itemize}[nosep,leftmargin=2em]
    \item
    Dropout and weight decay are turned off.

    \item
    To get representative training profiles for all models, the learning rates are tuned separately for \modelminik{1} and \modelminik{32} on validation sets using the usual metrics (i.e. RMSE or accuracy) as the guidance.
    The grid for learning rate tuning was: \mbox{\texttt{numpy.logspace(numpy.log10(1e-5), numpy.log10(5e-3), num=25)}}.

\end{itemize}

\subsection{Implementation details of \autoref{sec:analysis-selecting-submodels}}
\label{A:sec:impl-analysis-selecting-submodels}

\textbf{\modelgreedyheads.}
Here, we clarify the implementation details for \modelgreedyheads\ described in \autoref{sec:analysis-selecting-submodels}.
\modelgreedyheads\ is obtained from a trained \model\ by greedily selecting submodels from \model\ starting from the best one and stopping when two conditions are simultaneously true for the first time: (1) adding any new submodel does not improve the validation metric of the collective prediction; (2) the current validation metric is already better than that of the initial model with all $k$ submodels.
To clarify, during the greedy selection, the $i$-th submodel is considered to be better than the $j$-th submodel if adding the $i$-th submodel to the aggregated prediction leads to better validation metrics (i.e. it is \textit{not} the same as adding the submodel in the order of their individual validation metrics).

\subsection{Implementation details of \autoref{sec:analysis-k}}
\label{A:sec:impl-analysis-k}

\autoref{fig:d-k-ablation} shows the mean percentage improvements (see \autoref{A:sec:impl-metrics}) over MLP across $17$ datasets: all datasets except for Covertype from \autoref{A:tab:default-datasets}, and all datasets from TabReD \citep{rubachev2024tabred}.
We have used the dropout rate $0.1$ and tuned the learning rate separately for each value of $k$.
The score on each dataset is averaged over $5$ seeds.

\subsection{Non-linear embeddings for continuous features}
\label{A:sec:impl-feature-embeddings}

\textbf{Notation.}
We use the notation based on $\dagger$ and $\ddagger$ only for brevity.
Any other unambiguous notation can be used in future work.

\textbf{Updated piecewise-linear embeddings.}
We use a slightly different implementation of the piecewise-linear embeddings compared to \citet{gorishniy2022embeddings}.
Architecture-wise, our implementation corresponds to the ``Q-L'' and ``T-L'' variations from Table 2 in \citet{gorishniy2022embeddings} (we use the quantile-based bins for simplicity).
In practice, our implementation is significantly faster and uses a different parametrization and initialization.
See the source code for details.

\textbf{Other models.}
Since it is not feasible to test all combinations of backbones and embeddings, for baselines, we stick to the embeddings used in the original papers (applies to TabR \citep{gorishniy2023tabr}, ExcelFormer \citep{chen2023excelformer} and ModernNCA \citep{ye2024modern}).
For all models with feature embeddings (including TabM, MLP, TabR, ModernNCA, ExcelFormer), 
the embeddings-related details are commented in the corresponding sections below.

\subsection{\model}
\label{A:sec:impl-model}

\textbf{Feature embeddings.}
\modelminiemb\ and \modelemb\ are the versions of \model\ with non-linear feature embeddings.
\modelminiemb\ and \modelemb\ use the updated piecewise-linear feature embeddings mentioned in \autoref{A:sec:impl-feature-embeddings}.

\autoref{A:tab:tabm-space} provides the hyperparameter tuning spaces for \model\ and \modelmini.
\autoref{A:tab:tabm-emb-space} provides the hyperparameter tuning spaces for \modelemb\ and \modelminiemb.

\begin{table}[h!]
\centering
\caption{The hyperparameter tuning space for \model\ and \modelmini. Here, (B) = \{Covertype, Microsoft, \autoref{A:tab:tabred-datasets}\} and (A) contains all other datasets.}
{\renewcommand{\arraystretch}{1.2}
\begin{tabular}{lll}
    \toprule
    Parameter           & Distribution or Value\\
    \midrule
    $k$                   & $32$ \\
    \# layers           & $\mathrm{UniformInt}[1,5]$ \\
    Width (hidden size) & $\mathrm{UniformInt}[64,1024]$ \\
    Dropout rate        & $\{0.0, \mathrm{Uniform}[0.0,0.5]\}$ \\
    Learning rate       & $\mathrm{LogUniform}[1e\text{-}4, 5e\text{-}3]$ \\
    Weight decay        & $\{0, \mathrm{LogUniform}[1e\text{-}4, 1e\text{-}1]\}$ \\
    \midrule
    \# Tuning iterations & (A) 100  (B) 50 \\
    \bottomrule
\end{tabular}}
\label{A:tab:tabm-space}
\end{table}

\begin{table}[h!]
\centering
\caption{The hyperparameter tuning space for \modelminiemb\ and \modelemb. Here, (B) = \{Covertype, Microsoft, \autoref{A:tab:tabred-datasets}\} and (A) contains all other datasets.}
{\renewcommand{\arraystretch}{1.2}
\begin{tabular}{lll}
    \toprule
    Parameter           & Distribution or Value \\
    \midrule
    $k$                   & $32$ \\
    \# layers           & $\mathrm{UniformInt}[1,4]$ \\
    Width (hidden size) & $\mathrm{UniformInt}[64,1024]$ \\
    Dropout rate        & $\{0.0, \mathrm{Uniform}[0.0,0.5]\}$ \\
    \# PLE bins         & $\mathrm{UniformInt}[8, 32]$ \\
    Learning rate       & $\mathrm{LogUniform}[5e\text{-}5, 3e\text{-}3]$ \\
    Weight decay        & $\{0, \mathrm{LogUniform}[1e\text{-}4, 1e\text{-}1]\}$ \\
    \midrule
    \# Tuning iterations & (A) 100  (B) 50 \\
    \bottomrule
\end{tabular}}
\label{A:tab:tabm-emb-space}
\end{table}

\subsection{MLP}

\textbf{Feature embeddings.}
MLP\textsuperscript{$\dagger$} and MLP\textsuperscript{$\ddagger$} are the versions of MLP with non-linear feature embeddings.
MLP\textsuperscript{$\dagger$} uses the updated piecewise-linear embeddings mentioned in \autoref{A:sec:impl-feature-embeddings}.
MLP\textsuperscript{$\ddagger$} (also known as MLP-PLR) uses the periodic embeddings \citep{gorishniy2022embeddings}.
Technically, it is the \texttt{PeriodicEmbeddings} class from the \texttt{rtdl\_num\_embeddings} Python package.
We tested two variations: with \texttt{lite=False} and \texttt{lite=True}.
In the paper, only the former one is reported, but in the source code, the results for both are available.

\autoref{A:tab:mlp-space}, \autoref{A:tab:mlp-emb-space}, \autoref{A:tab:mlp-plr-space} provide the hyperparameter tuning spaces for MLP, MLP\textsuperscript{$\dagger$} and MLP\textsuperscript{$\ddagger$}, respectively.

\begin{table}[h!]
\centering
\caption{The hyperparameter tuning space for MLP.}
{\renewcommand{\arraystretch}{1.2}
\begin{tabular}{lll}
    \toprule
    Parameter           & Distribution \\
    \midrule
    \# layers           & $\mathrm{UniformInt}[1,6]$ \\
    Width (hidden size) & $\mathrm{UniformInt}[64,1024]$ \\
    Dropout rate        & $\{0.0, \mathrm{Uniform}[0.0,0.5]\}$ \\
    Learning rate       & $\mathrm{LogUniform}[3e\text{-}5, 1e\text{-}3]$ \\
    Weight decay        & $\{0, \mathrm{LogUniform}[1e\text{-}4, 1e\text{-}1]\}$ \\
    \midrule
    \# Tuning iterations & 100 \\
    \bottomrule
\end{tabular}}
\label{A:tab:mlp-space}
\end{table}

\begin{table}[h!]
\centering
\caption{The hyperparameter tuning space for $\mathrm{MLP}^{\dagger}$.}
{\renewcommand{\arraystretch}{1.2}
\begin{tabular}{lll}
    \toprule
    Parameter           & Distribution \\
    \midrule
    \# layers           & $\mathrm{UniformInt}[1,5]$ \\
    Width (hidden size) & $\mathrm{UniformInt}[64,1024]$ \\
    Dropout rate        & $\{0.0, \mathrm{Uniform}[0.0,0.5]\}$ \\
    Learning rate       & $\mathrm{LogUniform}[3e\text{-}5, 1e\text{-}3]$ \\
    Weight decay        & $\{0, \mathrm{LogUniform}[1e\text{-}4, 1e\text{-}1]\}$ \\
    \midrule
    d\_embedding      & $\mathrm{UniformInt}[8,32]$ \\
    n\_bins      & $\mathrm{UniformInt}[2,128]$ \\
    \# Tuning iterations & 100 \\
    \bottomrule
\end{tabular}}
\label{A:tab:mlp-emb-space}
\end{table}

\begin{table}[h!]
\centering
\caption{The hyperparameter tuning space for $\mathrm{MLP}^{\ddagger}$.}
{\renewcommand{\arraystretch}{1.2}
\begin{tabular}{lll}
    \toprule
    Parameter           & Distribution \\
    \midrule
    \# layers           & $\mathrm{UniformInt}[1,5]$ \\
    Width (hidden size) & $\mathrm{UniformInt}[64,1024]$ \\
    Dropout rate        & $\{0.0, \mathrm{Uniform}[0.0,0.5]\}$ \\
    Learning rate       & $\mathrm{LogUniform}[3e\text{-}5, 1e\text{-}3]$ \\
    Weight decay        & $\{0, \mathrm{LogUniform}[1e\text{-}4, 1e\text{-}1]\}$ \\
    \midrule
    n\_frequencies  & $\mathrm{UniformInt}[16,96]$ \\
    d\_embedding      & $\mathrm{UniformInt}[16,32]$ \\
    frequency\_init\_scale & $\mathrm{LogUniform}[1e\text{-}2, 1e\text{1}]$ \\
    \# Tuning iterations & 100 \\
    \bottomrule
\end{tabular}}
\label{A:tab:mlp-plr-space}
\end{table}

\subsection{TabR}

\textbf{Feature embeddings.}
TabR\textsuperscript{$\ddagger$} is the version of TabR with non-linear feature embeddings.
TabR\textsuperscript{$\ddagger$} uses the periodic embeddings \citep{gorishniy2022embeddings}, specifically, \texttt{PeriodicEmbeddings(lite=True)} from the \texttt{rtdl\_num\_embeddings} Python package on most datasets.
On the datasets from \autoref{A:tab:tabred-datasets}, TabR\textsuperscript{$\ddagger$} uses the \texttt{PeriodicEmbeddings(lite=True)} embeddings on the Sberbank Housing and Ecom Offers datasets, and \texttt{LinearReLUEmbeddings} on the rest (to fit the computations into the GPU memory, following the original TabR paper).

Since we follow the training and evaluation protocols from \cite{gorishniy2023tabr}, and TabR was proposed in \cite{gorishniy2023tabr}, we simply reuse the results for TabR.
More details can be found in Appendix.D from \cite{gorishniy2023tabr}.
When tuning TabR\textsuperscript{$\ddagger$} on the datasets from \autoref{A:tab:tabred-datasets}, we have used $25$ tuning iterations and the same tuning space as for TabR from \cite{rubachev2024tabred}.

\subsection{FT-Transformer}
We used the implementation from the "\texttt{rtdl\_revisiting\_models}" Python package. The results on datasets from \autoref{A:tab:tabred-datasets} were copied from \cite{rubachev2024tabred}, because the experiment setups are compatible.

\begin{table}[h!]
\centering
\caption{
    The hyperparameter tuning space for FT-Transformer \cite{gorishniy2021revisiting}.
    Here, (B) = \{Covertype, Microsoft\} and (A) contains all other datasets (except \autoref{A:tab:tabred-datasets}).
}
{\renewcommand{\arraystretch}{1.2}
\begin{tabular}{lll}
    \toprule
    Parameter           & Distribution or Value\\
    \midrule
    \# blocks           & $\mathrm{UniformInt}[1,4]$ \\
    $d_{token}$         & $\mathrm{UniformInt}[16,384]$ \\
    Attention dropout rate        & $\mathrm{Uniform}[0.0,0.5]$ \\
    FFN hidden dimension expansion rate       & $\mathrm{Uniform}[\nicefrac{2}{3},\nicefrac{8}{3}]$ \\
    FFN dropout rate & $\mathrm{Uniform}[0.0,0.5]$ \\
    Residual dropout rate & $\{0.0, \mathrm{Uniform}[0.0,0.2] \}$ \\
    Learning rate       & $\mathrm{LogUniform}[3e\text{-}5, 1e\text{-}3]$ \\
    Weight decay        & $\{0, \mathrm{LogUniform}[1e\text{-}4, 1e\text{-}1]\}$ \\
    \midrule
    \# Tuning iterations & (A) 100  (B) 50 \\
    \bottomrule
\end{tabular}}
\end{table}

\subsection{ModernNCA}

\textbf{Feature embeddings.}
We adapted the official implementation of \citet{ye2024modern}.
We used periodic embeddings \cite{gorishniy2022embeddings} (specifically, \texttt{PeriodicEmbeddings(lite=True)} from the \texttt{rtdl\_num\_embeddings} Python package) for ModernNCA\textsuperscript{$\ddagger$} and no embeddings for ModernNCA.
\autoref{A:tab:mnca-space} and \autoref{A:tab:mnca-emb-space} provides hyperparameter tuning spaces for each ModernNCA and ModernNCA\textsuperscript{$\ddagger$}.

\begin{table}[h!]
\centering
\caption{
    The hyperparameter tuning space for ModernNCA.
    Here, (C) = \{\autoref{A:tab:tabred-datasets}\}, (B) = \{Covertype, Microsoft\} and (A) contains all other datasets.
}
{\renewcommand{\arraystretch}{1.2}
\begin{tabular}{lll}
    \toprule
    Parameter           & Distribution \\
    \midrule
    \# blocks           & $\mathrm{UniformInt}[0, 2]$ \\
    $d_{block}$         & $\mathrm{UniformInt}[64,1024]$ \\
    dim         & $\mathrm{UniformInt}[64,1024]$ \\
    Dropout rate        & $\mathrm{Uniform}[0.0,0.5]$ \\
    Sample rate        & $\mathrm{Uniform}[0.05, 0.6]$ \\
    Learning rate       & $\mathrm{LogUniform}[1e\text{-}5, 1e\text{-}1]$ \\
    Weight decay        & $\{0, \mathrm{LogUniform}[1e\text{-}6, 1e\text{-}3]\}$ \\
    \midrule
    \# Tuning iterations & (A) 100  (B, C) 50 \\
    \bottomrule
\end{tabular}}
\label{A:tab:mnca-space}
\end{table}

\begin{table}[h!]
\centering
\caption{
    The hyperparameter tuning space for ModernNCA\textsuperscript{$\ddagger$}.
    Here, (C) = \{\autoref{A:tab:tabred-datasets}\}, (B) = \{Covertype, Microsoft\} and (A) contains all other datasets.
}
{\renewcommand{\arraystretch}{1.2}
\begin{tabular}{lll}
    \toprule
    Parameter           & Distribution \\
    \midrule
    \# blocks           & $\mathrm{UniformInt}[0, 2]$ \\
    $d_{block}$         & $\mathrm{UniformInt}[64,1024]$ \\
    dim         & $\mathrm{UniformInt}[64,1024]$ \\
    Dropout rate        & $\mathrm{Uniform}[0.0,0.5]$ \\
    Sample rate        & $\mathrm{Uniform}[0.05, 0.6]$ \\
    Learning rate       & $\mathrm{LogUniform}[1e\text{-}5, 1e\text{-}1]$ \\
    Weight decay        & $\{0, \mathrm{LogUniform}[1e\text{-}6, 1e\text{-}3]\}$ \\
    n\_frequencies & $\mathrm{UniformInt}[16, 96]$ \\
    d\_embedding & $\mathrm{UniformInt}[16, 32]$ \\
    frequency\_init\_scale & $\mathrm{LogUniform}[0.01, 10]$ \\
    \midrule
    \# Tuning iterations & (A) 100  (B, C) 50 \\
    \bottomrule
\end{tabular}}
\label{A:tab:mnca-emb-space}
\end{table}

\subsection{T2G-Former}
We adapted the implementation and hyperparameters of \cite{yan2023t2g} from the official repository\footnote{https://github.com/jyansir/t2g-former}. \autoref{A:tab:t2g-space} provides hyperparameter tuning space.

\begin{table}[h!]
\centering
\caption{The hyperparameter tuning space for T2G-Former \cite{yan2023t2g}. Here, (C) = \{\autoref{A:tab:tabred-datasets}\}, (B) = \{Covertype, Microsoft\} and (A) contains all other datasets. Also, we used $50$ tuning iterations on some datasets from \cite{grinsztajn2022why}.}
{\renewcommand{\arraystretch}{1.2}
\begin{tabular}{lll}
    \toprule
    Parameter           & Distribution or Value\\
    \midrule
    \# blocks           & (A) $\mathrm{UniformInt}[3,4]$ (B, C) $\mathrm{UniformInt}[1,3]$\\
    $d_{token}$         & $\mathrm{UniformInt}[64,512]$ \\
    Attention dropout rate        & $\mathrm{Uniform}[0.0,0.5]$ \\
    FFN hidden dimension expansion rate       & (A, B) $\mathrm{Uniform}[\nicefrac{2}{3},\nicefrac{8}{3}]$ (C) $4/3$ \\
    FFN dropout rate & $\mathrm{Uniform}[0.0,0.5]$ \\
    Residual dropout rate & $\{0.0, \mathrm{Uniform}[0.0,0.2] \}$ \\
    Learning rate       & $\mathrm{LogUniform}[3e\text{-}5, 1e\text{-}3]$ \\
    Col. Learning rate       & $\mathrm{LogUniform}[5e\text{-}3, 5e\text{-}2]$ \\
    Weight decay        & $\{0, \mathrm{LogUniform}[1e\text{-}6, 1e\text{-}1]\}$ \\
    \midrule
    \# Tuning iterations & (A) 100  (B) 50 (C) 25 \\
    \bottomrule
\end{tabular}}
\label{A:tab:t2g-space}
\end{table}

\subsection{SAINT}
We completely adapted hyperparameters and protocol from \cite{gorishniy2023tabr} to evaluate SAINT on \cite{grinsztajn2022why} benchmark. Results on datasets from \autoref{A:tab:default-datasets} were directly taken from \cite{gorishniy2023tabr}. Additional details can be found in Appendix.D from \cite{gorishniy2023tabr}. We have used a default configuration on big datasets due to the very high cost of tuning (see \autoref{A:tab:saint-hp}).

\begin{table}[h!]
\centering
\caption{The default hyperparameters for SAINT \citep{somepalli2021saint} on datasets from \cite{rubachev2024tabred}.}
\label{A:tab:saint-hp}
{\renewcommand{\arraystretch}{1.2}
\begin{tabular}{lll}
    \toprule
    Parameter           & Value \\
    \midrule
    depth           & $2$\\
    $d_{token}$ & $32$ \\
    $n_{heads}$         & $4$ \\
    $d_{head}$         & $8$ \\
    Attention dropout rate        & $0.1$ \\
    FFN hidden dimension expansion rate       & $1$ \\
    FFN dropout rate & $0.8$ \\
    
    Learning rate       & $1e\text{-}4$ \\
    Weight decay        & $1e\text{-}2$ \\
    \bottomrule
\end{tabular}}
\end{table}

\subsection{Excelformer}

\textbf{Feature embeddings.}
ExcelFormer \citep{chen2023excelformer} uses custom non-linear feature embeddings based on a GLU-style activation, see the original paper for details.

We adapted the implementation and hyperparameters of \cite{chen2023excelformer} from the official repository\footnote{https://github.com/WhatAShot/ExcelFormer}.
For a fair comparison with other models, we did not use the augmentation techniques from the paper in our experiments.
See \autoref{A:tab:excel-space}.

\begin{table}[h!]
\centering
\caption{The hyperparameter tuning space for Excelformer \cite{chen2023excelformer}. Here, (D) = \{Homecredit, Maps Routing\}, (C) = \{\autoref{A:tab:tabred-datasets} w/o (D)\}, (B) = \{Covertype, Microsoft\} and (A) contains all other datasets.}
{\renewcommand{\arraystretch}{1.2}
\begin{tabular}{lll}
    \toprule
    Parameter           & Distribution or Value \\
    \midrule
    \# blocks           & (A, B) $\mathrm{UniformInt}[2,5]$ (C) $\mathrm{UniformInt}[2,4]$ (D) $\mathrm{UniformInt}[1,3]$ \\
    $d_{token}$         & (A, B) $\{32, 64, 128, 256\}$ (C) $\{16, 32, 64\}$ (D) $\{4, 8, 16, 32\}$ \\
    $n_{heads}$         & (A,B) $\{4, 8, 16, 32\}$ (C) $\{4, 8, 16\}$ (D) $4$ \\
    Attention dropout rate        & $0.3$ \\
    FFN dropout rate & $0.0$ \\
    Residual dropout rate & $\mathrm{Uniform}[0.0,0.5]$ \\
    Learning rate       & $\mathrm{LogUniform}[3e\text{-}5, 1e\text{-}3]$ \\
    Weight decay        & $\{0, \mathrm{LogUniform}[1e\text{-}4, 1e\text{-}1]\}$ \\
    \midrule
    \# Tuning iterations & (A) 100  (B) 50 (C, D) 25 \\
    \bottomrule
\end{tabular}}
\label{A:tab:excel-space}
\end{table}

\subsection{CatBoost, XGBoost and LightGBM}
Since our setup is directly taken from \cite{gorishniy2023tabr}, we simply reused their results for GBDTs from the official repository\footnote{https://github.com/yandex-research/tabular-dl-tabr}. 
Importantly, in a series of preliminary experiments, we confirmed that those results are reproducible in our instance of their setup.
The details can be found in Appendix.D from \cite{gorishniy2023tabr}. Results on datasets from \autoref{A:tab:tabred-datasets} were copied from the paper \citep{rubachev2024tabred}.

\subsection{AutoInt}

We used an implementation from \cite{gorishniy2021revisiting} which is an adapted official implementation\footnote{https://github.com/shichence/AutoInt}.

\begin{table}[h!]
\centering
\caption{The hyperparameter tuning space for AutoInt \citep{song2019autoint}. Here, (B) = \{Covertype, Microsoft\} and (A) contains all other datasets.}
{\renewcommand{\arraystretch}{1.2}
\begin{tabular}{lll}
    \toprule
    Parameter           & Distribution \\
    \midrule
    \# blocks           & $\mathrm{UniformInt}[1,6]$ \\
    $d_{token}$         & $\mathrm{UniformInt}[8,64]$ \\
    $n_{heads}$         & 2 \\
    Attention dropout rate        & $\{0, \mathrm{Uniform}[0.0,0.5]\}$ \\
    Embedding dropout rate & $\{0, \mathrm{Uniform}[0.0,0.5]\}$ \\
    Learning rate       & $\mathrm{LogUniform}[3e\text{-}5, 1e\text{-}3]$ \\
    Weight decay        & $\{0, \mathrm{LogUniform}[1e\text{-}4, 1e\text{-}1]\}$ \\
    \midrule
    \# Tuning iterations & (A) 100  (B) 50 \\
    \bottomrule
\end{tabular}}
\vspace{1em}
\end{table}

\subsubsection{TabPFN}
Since TabPFN accepts only less than 10K training samples we use different subsamples of size 10K for different random seeds. Also, TabPFN is not applicable to regressions and datasets with more than $100$ features.

\newpage

\section{Per-dataset results with standard deviations}
\label{A:sec:per-dataset-results}
\centering

\newcommand{\topalign}[1]{%
\vtop{\vskip 0pt #1}}

\begin{longtable}{p{0.5\textwidth}p{0.5\textwidth}}
\caption{
Extended results for the main benchmark.
Results are grouped by datasets.
One ensemble consists of five models trained independently under different random seeds.
}\\
\end{longtable}

\begin{longtable}{p{0.5\textwidth}p{0.5\textwidth}}
\caption{Extended results for \cite{grinsztajn2022why} benchmark. Results are grouped by datasets. One ensemble consists of five models trained independently with different random seeds.}\\
\end{longtable}

\begin{longtable}{p{0.5\textwidth}p{0.5\textwidth}}
\caption{Extended results for TabReD \cite{rubachev2024tabred} benchmark. Results are grouped by datasets. One ensemble consists of five models trained independently under different random seeds.}\\
\end{longtable}

\end{document}